\documentclass{article}

\usepackage[final]{neurips_2023}

\usepackage[utf8]{inputenc} 
\usepackage[T1]{fontenc}    
\usepackage{hyperref}       
\hypersetup{
    colorlinks = true,
    citecolor = blue,
    linkcolor = magenta}
\usepackage{url}            
\usepackage{booktabs}       
\usepackage{amsfonts}       
\usepackage{nicefrac}       
\usepackage{microtype}      
\usepackage{xcolor}         
\usepackage{amsmath}
\usepackage{graphicx}
\usepackage{tabularx}
\usepackage{hyperref}
\usepackage[utf8]{inputenc}
\usepackage{natbib}
\usepackage{xcolor}
\usepackage{ifthen}
\usepackage[normalem]{ulem}
\usepackage{svg}
\usepackage{subfiles}
\usepackage{xspace}
\usepackage{wrapfig}
\usepackage{dialogue}
\usepackage[most]{tcolorbox}
\usepackage{subcaption}
\usepackage{booktabs}
\usepackage{multirow}

\newcommand{\1}{\mathbb{I}} 

\ifthenelse{\isundefined{\definition}}{\newtheorem{definition}{Definition}}{}
\ifthenelse{\isundefined{\assumption}}{\newtheorem{assumption}{Assumption}}{}
\ifthenelse{\isundefined{\hypothesis}}{\newtheorem{hypothesis}{Hypothesis}}{}
\ifthenelse{\isundefined{\proposition}}{\newtheorem{proposition}{Proposition}}{}
\ifthenelse{\isundefined{\theorem}}{\newtheorem{theorem}{Theorem}}{}
\ifthenelse{\isundefined{\lemma}}{\newtheorem{lemma}{Lemma}}{}
\ifthenelse{\isundefined{\corollary}}{\newtheorem{corollary}{Corollary}}{}
\ifthenelse{\isundefined{\alg}}{\newtheorem{alg}{Algorithm}}{}
\ifthenelse{\isundefined{\example}}{\newtheorem{example}{Example}}{}

\def\Snospace~{\S{}}

\newcommand\eg{e.g.\xspace}
\newcommand\ie{i.e.\xspace}

\newcommand\system{system\xspace}

\newcommand\systemic{systemic\xspace}
\newcommand\systemlevel{ecosystem-level\xspace}
\newcommand\Systemlevel{Ecosystem-level\xspace}
\newcommand{\baseline}{baseline\xspace}

\newcommand\profilepolarization{homogeneous outcomes\xspace}
\newcommand\Profilepolarization{Homogeneous outcomes\xspace}
\newcommand\polarization{homogeneity\xspace}

\newcommand\polarized{homogenous\xspace}
\newcommand\polar{homogeneous\xspace}
\newcommand\Profiles{Outcomes\xspace}
\newcommand\polarizedprofiles{homogenous outcomes\xspace}
\newcommand\PolarizedProfiles{Homogenous Outcomes\xspace}
\newcommand\polaroutcomes{homogeneous outcomes\xspace}

\newcommand\outcomeprofile{outcome profile\xspace}
\newcommand\outcomeprofiles{outcome profiles\xspace}
\newcommand\Outcomeprofile{Outcome profile\xspace}
\newcommand\Outcomeprofiles{Outcome profiles\xspace}
\newcommand\slightlyantipolarized{least homogeneous\xspace}
\newcommand\unpolarized{heterogeneous\xspace}
\newcommand\antipolarization{heterogeneity\xspace}

\newcommand\pl[1]{\textcolor{red}{[PL: #1]}}
\newcommand\rb[1]{\textcolor{pink}{[RB: #1]}}
\newcommand\ct[1]{\textcolor{blue}{[CT: #1]}}
 \newcommand\kc[1]{\textcolor{violet}{[KC: #1]}}
 \newcommand\shb[1]{\textcolor{orange}{[SHB: #1]}}
 \newcommand\tldr[1]{\textcolor{gray}{\textit{[#1]}}}

 \renewcommand\pl[1]{}
 \renewcommand\rb[1]{}
 \renewcommand\ct[1]{}
 \renewcommand\kc[1]{}
 \renewcommand\shb[1]{}
 \renewcommand\tldr[1]{}

\newcommand{\hapi}{\textbf{HAPI}\xspace}

\newcommand{\digit}{\textsc{digit}\xspace} 
\newcommand{\fluent}{\textsc{fluent}\xspace} 
\newcommand{\amnist}{\textsc{amnist}\xspace} 
\newcommand{\command}{\textsc{command}\xspace} 

\newcommand{\yelp}{\textsc{yelp}\xspace} 
\newcommand{\imdb}{\textsc{imdb}\xspace} 
\newcommand{\shop}{\textsc{shop}\xspace} 
\newcommand{\waimai}{\textsc{waimai}\xspace} 

\newcommand{\ferplus}{\textsc{fer+}\xspace} 
\newcommand{\afnet}{\textsc{afnet}\xspace} 
\newcommand{\expw}{\textsc{expw}\xspace} 
\newcommand{\rafdb}{\textsc{rafdb}\xspace} 

\newcommand{\ddi}{\textbf{DDI}\xspace}

\newcommand{\himproved}{$h_\text{imp}$\xspace}

\newcommand{\githubURL}{\url{https://github.com/rishibommasani/EcosystemLevelAnalysis}}
\newcommand\tldrDone[1]{}

\usepackage[disable,textsize=tiny]{todonotes}

\title{
Ecosystem-level Analysis of Deployed Machine Learning Reveals Homogeneous Outcomes}

\author{
Connor Toups\thanks{Equal contribution.} \\
Stanford University\\
\And
Rishi Bommasani\footnotemark[1]~~\thanks{Corresponding author: \url{nlprishi@stanford.edu}.} \\
Stanford University\\
\AND 
Kathleen A. Creel \\
Northeastern University\\
\And 
Sarah H. Bana \\
Chapman University\\
\And 
Dan Jurafsky \\
Stanford University\\
\And 
Percy Liang \\
Stanford University\\
}

\begin{document}
\maketitle
\setcounter{footnote}{0}
\begin{abstract}
Machine learning is traditionally studied at the model level: researchers measure and improve the accuracy, robustness, bias, efficiency, and other dimensions of specific models.
In practice, however, the societal impact of any machine learning model depends on the context into which it is deployed. 
To capture this, we introduce \textit{\systemlevel analysis}: rather than analyzing a single model, we consider the collection of models that are deployed in a given context.
For example, \systemlevel analysis in hiring recognizes that a job candidate's outcomes are determined not only by a single hiring algorithm or firm but instead by the collective decisions of all the firms to which the candidate applied.
Across three modalities (text, images, speech) and eleven datasets, we establish a clear trend: deployed machine learning is prone to \textit{systemic failure}, meaning some users are exclusively misclassified by all models available. 
Even when individual models improve over time, we find these improvements rarely reduce the prevalence of systemic failure. 
Instead, the benefits of these improvements predominantly accrue to individuals who are already correctly classified by other models. 
In light of these trends, we analyze medical imaging for dermatology, a setting where the costs of systemic failure are especially high. 
While traditional analyses reveal that both models and humans exhibit racial performance disparities, \systemlevel analysis reveals new forms of racial disparity in model predictions that do not present in human predictions.
These examples demonstrate that \systemlevel analysis has unique strengths in characterizing the societal impact of machine learning.\footnote{All code is available at \githubURL.}
\end{abstract}
\section{Introduction}
\label{sec:introduction}

Machine learning (ML) is pervasively deployed. Systems based on ML mediate our communication and healthcare,
influence where we shop or what we eat,
and allocate opportunities like loans and jobs.
Research on the societal impact of ML typically focuses on the behavior of individual models. 
If we center people, however, we recognize that the impact of ML on our lives depends on the aggregate result of our many interactions with ML models.


\begin{figure}
    \centering
    \begin{subfigure}[b]{0.4\textwidth}
        \centering
        \includegraphics[width=\textwidth]{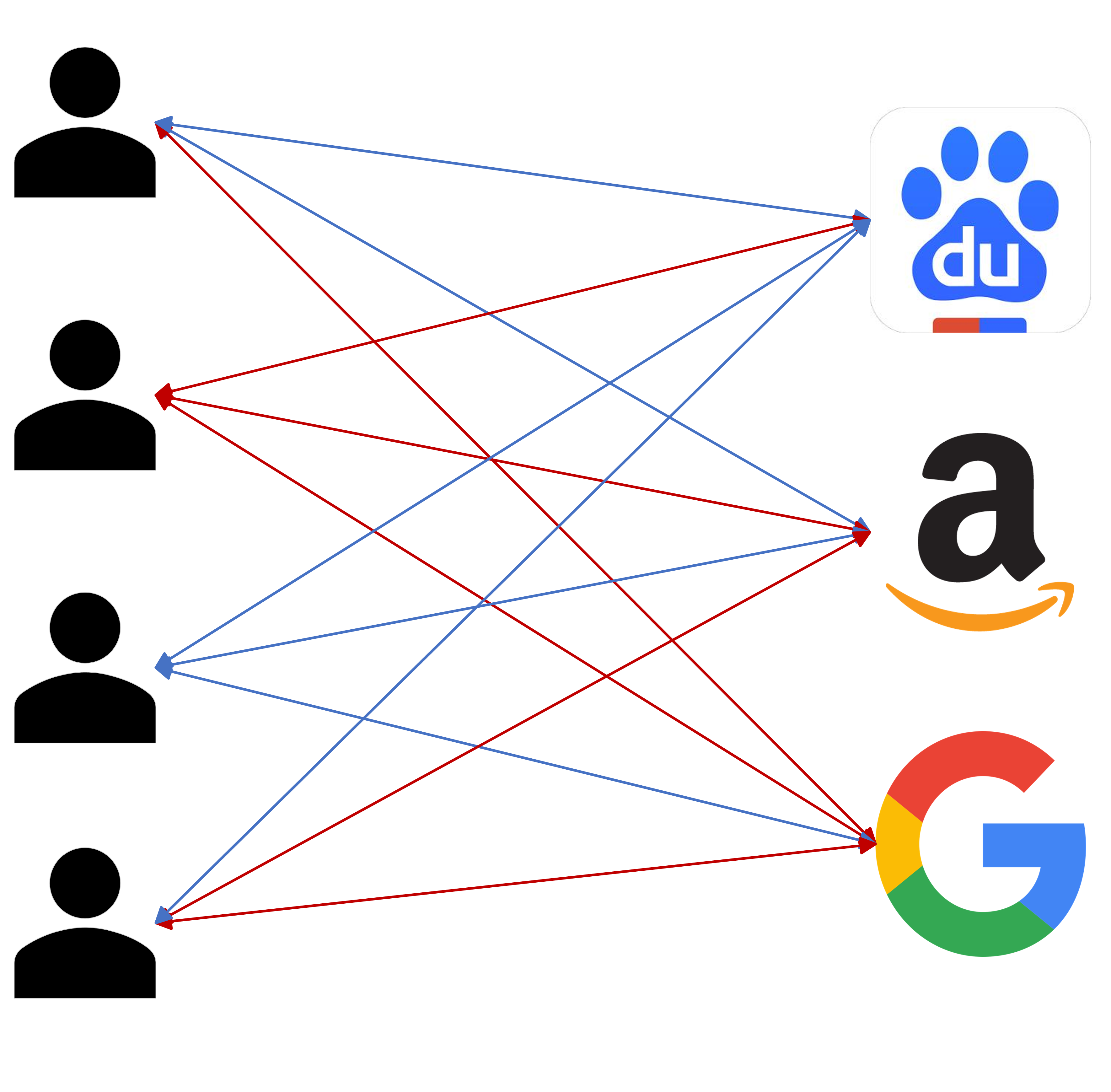}
    \end{subfigure}%
    \hspace{3em}%
    \begin{subfigure}[b]{0.4\textwidth}
        \centering
        \includegraphics[width=\textwidth]{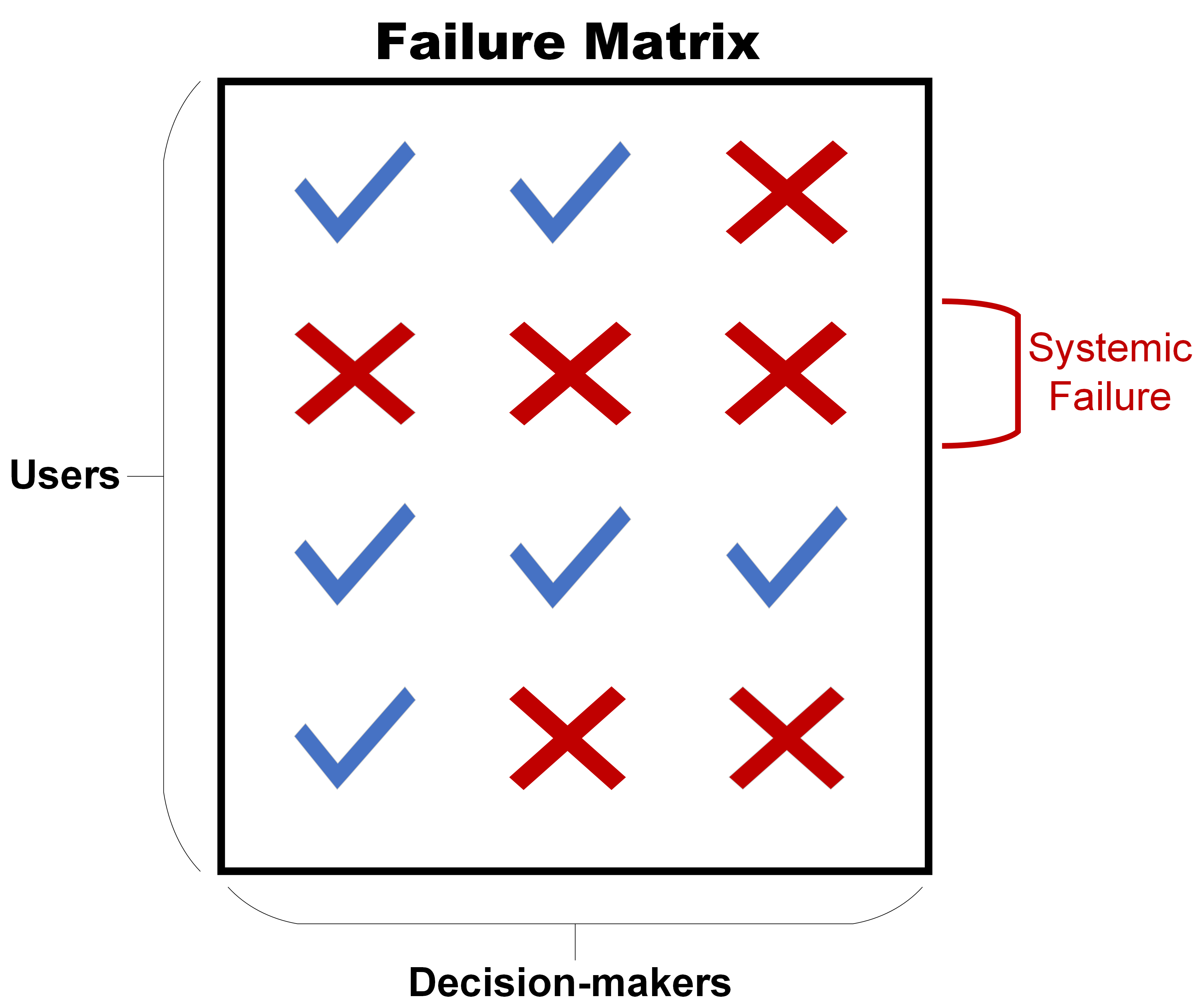}
    \end{subfigure}%
    \caption{\textbf{\Systemlevel analysis.} 
    Individuals interact with decision-makers (\textit{left)}, receiving outcomes that constitute the failure matrix (\textit{right}).
    }
    \label{fig:fig1}
\end{figure}

In this work, we introduce \textit{\systemlevel analysis} to better characterize the societal impact of machine learning on people.
Our insight is that when a ML model is deployed, the impact on users depends not only on its behavior but also on the behavior of other models and decision-makers  (left of \autoref{fig:fig1}).
For example, the decision of a single hiring algorithm to reject or accept a candidate does not determine whether or not the candidate secures a job; the outcome of her search depends on the decisions made by all the firms to which she applied.  
Likewise, in selecting consumer products like voice assistants, users choose from options such as Amazon Alexa, Apple Siri, or Google Assistant. From the user's perspective, what is important is that at least one product works.

In both settings, there is a significant difference from the user's perspective between \textit{systemic failure}, in which \textit{zero} systems correctly evaluate them or work for them, and any other state.  The difference in marginal utility between zero acceptances and one acceptance is typically much higher than the difference between one acceptance and two acceptances. This non-linearity is not captured by average error metrics. 
For example, imagine that three companies are hiring and there are ten great candidates. All three companies wrongly reject the same two great candidates for a false negative error rate of 20\%. 
Now imagine that each company wrongly rejects \textit{different} candidates.  The second decision ecosystem has the same false negative error rate, 20\%, but no systemic failures or jobless candidates.   
\pl{this is an important point, would explain more:there is a big difference between 0 working and nonzero working,the marginal utility of 0 to 1 is much higher than 1 to 2;this non-linearity is not captured by average error metrics;you could even have a simple example where two systems have the same average error rate, but one has lots of systemic failures}\kc{Addressed.}

\Systemlevel analysis is a methodology that centers on the \textit{failure matrix} ($F$;  right of \autoref{fig:fig1}): $F$ encapsulates the outcomes individuals receive from all decision-makers.
Of special interest are \textit{systemic failures}: exclusively negative outcomes for individuals such as misclassifications or rejections from all decision-makers \citep{bommasani2022picking}.
To establish general trends in deployed machine learning, we draw upon a large-scale audit \citep[\hapi;][]{chen2022hapi} that spans three modalities (images, speech, text), three commercial systems per modality, and eleven datasets overall. Because \hapi contains predictions from some of the largest commercial ML providers -- including Google, Microsoft, and Amazon -- and the models evaluated are deployed models that real users interact with, evaluating \hapi has real-world implications. 

Across all settings, \systemlevel analysis reveals a consistent pattern of \textit{\profilepolarization}. 
In each of the \hapi datasets, many instances are classified correctly by all three commercial systems and many instances are classified incorrectly by all three systems.  The pattern of outcomes across the decision ecosystem is \textit{\polarized} if the rates of systemic failure and consistent classification success significantly exceed the rate predicted by independent instance-level behavior. Since the  commercial systems analyzed in this dataset are popular and widely used, 
being failed by all systems in the dataset has societally meaningful consequences.  

\Systemlevel analysis enriches our understanding not only of the status quo, but also of how models change over time.
In particular, it allows us to ask, when individual models improve, how do \systemlevel outcomes change?
Since \hapi tracks the performance of the same systems over a three-year period, we consider all cases where at least one of the commercial systems improves.
For example, Amazon's sentiment analysis API reduced its error rate on the \waimai dataset by 2.5\% from 2020 to 2021; however, this improvement did not decrease the systemic failure rate at all. Precisely 0 out of the model's 303 improvements are on instances on which all other models had failed.
These findings generalize across all cases: on average, just 10\% of the instance-level improvement of a single commercial system occurs on instances misclassified by all other models.  This is true even though \systemic failures account for 27\% of the instances on which the models \textit{could have} improved. Thus most model improvements do not significantly reduce \systemic failures. 

To build on these trends, we study medical imaging, a setting chosen because the costs to individuals of \systemic failure of medical imaging classification are especially high. 
We compare outcomes from prominent dermatology models and board-certified dermatologists on the \ddi dataset \citep{daneshjou2022disparities}: both models and humans demonstrate \profilepolarization, though human outcomes are more \polarized. 
Given established racial disparities in medicine for both models and humans, fairness analyses in prior work show that both humans and models consistently perform worse for darker skin tones (\eg \citet{daneshjou2022disparities} show lower ROC-AUC on \ddi).
\Systemlevel analysis surfaces new forms of racial disparity in models that do not present in humans: models are more \textit{\polarized} when evaluating images with darker skin tones, meaning that all systems agree in their correct or incorrect classification, whereas human \polarization is consistent across skin tones.
\pl{unclear what polarization is at this point but current writing presupposes that the reader should know}\kc{Addressed by changing it to homogeneity and adding a macro for all uses.}

Our work contributes to a growing literature on the \textit{homogeneous outcomes} of modern machine learning \citep{ajunwa2019paradox, Engler2021, creel2022, bommasani2022picking,  fishman2022should, wang2023overcoming, jain2023pluralism}.
While prior work conceptualizes these phenomena, our work introduces new methodology to study these problems and provides concrete findings for a range of ML deployments spanning natural language processing, computer vision, speech, and medical imaging.
Further, by centering individuals, we complement established group-centric methods \citep[][]{barocas2016disparate, buolamwini2018gender, koenecke2020racial}, unveiling new forms of racial disparity.
\Systemlevel analysis builds on this existing work, providing a new tool that contributes to holistic evaluations of the societal impact of machine learning.

Developing better methodologies for \systemlevel analysis of deployed machine learning systems is important for two reasons.  First, 
systemic failures in socially consequential domains could exclude people from accessing goods such as jobs, welfare benefits, or correct diagnoses.
Individuals who are failed by only one model can gain informal redress by switching to another model, for example by seeking second doctor's opinion or switching banks. Individuals failed by \textit{all} models cannot.   \pl{this seems vague; basically you want to say that such people risk being excluded from society, not just excluded from one system; if you have a system-level failure, you might choose another system, but if you have a ecosystem-level failure, there is no recourse [oh you say this at the end of the paragraph; I'd promote this]}\kc{addressed} 
Socially consequential systemic failures can happen due to reliance on APIs, such as image recognition APIs used to identify cancers, speech recognition APIs used to verify individuals for banking, or facial recognition APIs used to unlock devices. Systemic failures can also occur in algorithmic decision-making systems such as those used for hiring, lending, and criminal justice. The social importance of avoiding systemic failures in all of these systems is clear. 

Second, as decision-makers become more likely to rely on the same or similar algorithms to make decisions  \citep{kleinberg2021monoculture} or to use the same or similar components in building their decision pipelines  \citep{bommasani2022picking}, we believe that the prevalence of systemic failures could increase. Measuring systemic failures as they arise with the tools we present in this paper will expand our understanding of their prevalence and likely causes.

\pl{at some point, we should discuss expected baselines, probably in the discussion section - the fact that some individuals are 'easier' and some individuals are 'harder'}\kc{This is currently addressed later but we can add a reference here in the intro if you would like.}
\section{\Systemlevel Analysis}
\label{sec:preliminaries}
How individuals interact with deployed machine learning models determines ML's impact on their lives.
In some contexts, individuals routinely interact with multiple ML models.
For example, when a candidate applies to jobs, they typically apply to several firms. 
The decision each company makes to accept or reject the candidate may be mediated by a single hiring algorithm. 
In other contexts, individuals select a single model from a set of options.
For example, when a consumer purchases a voice assistant, they typically choose between several options (\eg Amazon Alexa, Google Assistant, Apple Siri) to purchase a single product (\eg Amazon Alexa).
Centering people reveals a simple but critical insight: exclusively receiving negative outcomes, as when individuals are rejected from every job or unable to use every voice assistant, has more severe consequences than receiving even one positive outcome. 
\pl{clearly zero is 'worst', but this is not an insight; the insight is that there is special significance to a zero}\kc{addressed}

\subsection{Definition}
\label{sec:system-level}
Recognizing how ML is deployed, we introduce \textit{\systemlevel analysis} as a methodology for characterizing ML's cumulative impacts on individuals.
Consider $N$ individuals that do, or could, interact with $k$ decision-makers that apply $\hat{y}$ labels according to their decision-making processes $h_1, \dots, h_k$.
Individual $i$ is associated with input $x_i$, label $y_i$, and receives the label $\hat{y}_i^j = h_j(x_i)$ from decision-maker $j$.
\pl{label and decision are different types - can't really compare them; I'd probably stick with label for simplicity; otherwise, if you have a hiring rejection decision, what is the 'label'? I think this paper is mostly about prediction and accuracies}\kc{Addressed here; may need to be addressed elsewhere}

\paragraph{\Profiles.}
Define the \textit{failure matrix} $F \in \{0, 1\}^{N \times k}$ such that $F[i, j] = \1 \left[\hat{y}_i^j \neq y_i\right]$.
The \textit{failure outcome profiles} $\mathbf{f}_i$ for individual $i$, which we refer to as the \outcomeprofile for brevity, denotes $F[i, :]$.
The \textit{failure rate} $\bar{f_j}$ for decision-maker $j$ is $\bar{f_j} = \frac{\sum_{i=1}^N F[i, j]}{N}$ (\ie the empirical classification error in classification).
For consistency, we order the entries of decision-makers (and, thereby, the columns of the failure matrix) in order of ascending failure rate: $F[:, 1]$ is the \outcomeprofile associated with the decision-maker with the fewest failures and $F[:, k]$ is the \outcomeprofile associated with the decision-maker with the most failures.
The failure matrix is the central object in \systemlevel analysis (see \autoref{fig:fig1}). 

\paragraph{Systemic Failures.}
Individual $i$ experiences \textit{systemic failure} if they exclusively experience failure across the domain of interest: $F[i, :] = [1, \dots, 1]$.
Not only are systemic failures the worst possible outcomes, but they also often result in additional harms. If an individual applying to jobs is rejected everywhere, they may be unemployed.
If no commercial voice assistant can recognize an individual's voice, they may be fully locked out of accessing a class of technology.
In our \systemlevel analysis, we focus on systemic failures as a consequential subset of the broader class of \textit{homogeneous outcomes} \citep{bommasani2022picking}.

\section{Homogeneous Outcomes in Commercial ML APIs (HAPI)}
\label{sec:hapi-polarization}
To establish general trends made visible through \systemlevel analysis, we draw upon a large-scale three-year audit of commercial ML APIs \citep[\hapi;][]{chen2022hapi} to study the behavior of deployed ML systems across three modalities, eleven datasets, and nine commercial systems.

\subsection{Data}
\label{sec:hapi-data}

\citet{chen2022hapi} audit commercial ML APIs, tracking predictions across these APIs when evaluated on the same eleven standard datasets over a period of three years (2020 -- 2022).
We consider ML APIs spanning three modalities (text, images, speech), where each modality is associated with a task (SA: sentiment analysis, FER: facial emotion recognition, SCR: spoken command recognition) and 3 APIs per modality (\eg IBM, Google, Microsoft for spoken command recognition). The models evaluated are from Google (SA, SCR, FER), Microsoft (SCR, FER), Amazon (SA), IBM (SCR), Baidu (SA), and Face++ (FER).
Additionally, each modality is associated with three to four datasets, amounting to eleven datasets total; further details are deferred to the supplement.

To situate our notation, consider the \digit dataset for spoken command recognition and the associated APIs (IBM, Google, Microsoft). 
For each instance (i.e. image) $x_i$ in \digit, the \outcomeprofile $\textbf{f}_i \in \{0, 1\}^3$ is the vector of outcomes.
The entries are ordered by ascending model failure rate: $F[:, 1]$ corresponds to the most accurate model (Microsoft) and $F[:, 3]$ corresponds to the least accurate model (Google).
\pl{before rows $i$ were individuals; but presumably in these datasets, they are not individuals but instances; either way, worth clarifying, because this affects our intuitions / interpretations of what a systemic failure really means} \shb{changed datapoint to instance (i.e. image)}

\paragraph{Descriptive statistics.}
\begin{table*}[htp]
\resizebox{\textwidth}{!}{
\begin{tabular}{l|cccc|ccc|cccc}
\toprule
 & \multicolumn{4}{c}{\textbf{Facial emotion recognition}} & \multicolumn{3}{c}{\textbf{Spoken command recognition}} & \multicolumn{4}{c}{\textbf{Sentiment analysis}} \\
 &   \rafdb &    \afnet &    \expw & \ferplus &  \fluent &  \digit &  \amnist &    \shop &    \yelp &    \imdb & \waimai \\
\midrule
\textbf{Dataset size} &  15.3k &  287.4k &  31.5k &    6.4k &  30.0k &   2.0k &  30.0k &  62.8k &  20.0k &  25.0k &  12.0k \\
\textbf{Number of classes} &       7 &        7 &       7 &       7 &      31 &     10 &      10 &       2 &       2 &       2 &       2 \\
\midrule
\textbf{$h_1$ failure rate} (\ie error) &   0.283 &    0.277 &   0.272 &   0.156 &   0.019 &  0.217 &   0.015 &   0.078 &   0.043 &   0.136 &    0.110 \\
\textbf{$h_2$ failure rate} (\ie error) &   0.343 &    0.317 &   0.348 &   0.316 &   0.025 &  0.259 &   0.015 &   0.095 &   0.111 &   0.219 &   0.151 \\
\textbf{$h_3$ failure rate} (\ie error) &   0.388 &    0.359 &   0.378 &   0.323 &   0.081 &  0.472 &   0.043 &   0.122 &   0.486 &   0.484 &   0.181 \\
\textbf{Systemic failure rate} &   0.152 &    0.178 &   0.181 &   0.066 &    0.01 &  0.129 &   0.002 &   0.039 &   0.021 &   0.043 &   0.065 \\
\bottomrule
\end{tabular}}
\caption{\textbf{Basic statistics on \hapi datasets} including the (observed) systemic failure rate (\ie fraction of instances misclassified by all models).}
\label{tab:basic-stats}
\end{table*}

To build general understanding of model performance in \hapi, we provide basic descriptive statistics (\autoref{tab:basic-stats}).
For most datasets, all APIs achieve accuracies within 5--10\% of each other (exceptions include \digit, \yelp, \imdb). 
Interestingly, we often find the systemic failure rate is roughly half the failure rate of the most accurate model $h_1$.
\pl{why is this last sentence relevant? failure rate of $h_3$ is an even tighter upper bound}\kc{Commented out the final sentence.}

\subsection{\Systemlevel Behavior}

In order for a measure of systemic failure to be useful, it must be (i) meaningful and (ii) comparable across systems. A challenge to the meaningfulness of any proposed metric is that 
systemic failures occur more often in an ecosystem with many inaccurate models.  A metric for systemic failure that primarily communicated the aggregate error rates of models in the ecosystem would not be meaningful as an independent metric.  It also would not support goal (ii) because we could not compare the rates of systemic failure across ecosystems with varying model accuracies. It would be difficult to identify a system with a `large' rate of systemic failure because 
the systemic failure properties would be swamped by the error rates of the models in the ecosystem. Therefore, achieving meaningfulness and comparability requires the metric to incorporate error correction.

Assuming model independence is a helpful baseline because it adjusts for model error rates without making assumptions about the correlation between models in the ecosystem. To avoid assumptions and for the sake of simplicity, therefore, we juxtapose the \textit{observed} behavior with a simple theoretical model in which we assume models fail independently of each other.
Under this assumption, the distribution of the \textit{\baseline} number of model failures $t \in \{0, \dots, k\}$ follows a Poisson-Binomial distribution parameterized by their failure rates (\autoref{p_baseline}). 
\pl{I'd flip things to first justify why you need a baseline (most of this paragraph), and then in a separate paragraph, say we propose the independent baseline due to simplicity}\kc{addressed}

The baseline of independence also means that our metric does not attempt to quantify whether it is ``reasonable'' that the models all fail on some instances. For example, some instances might be harder (or easier) than others, making it more likely that all models will fail (or succeed) to classify that instance.  However, ``hardness'' is observer-relative.  What is hard for one class of models might be easy for another class of model, and likewise with humans with particular capabilities or training.  Therefore 'correcting' the metric to account for hardness would relativize the metric to the group of humans or class of models for whom that instance is hard.  We choose independence as a baseline to be neutral on this point.  However, we depart from independence in Appendix \autoref{sec:expressive-model}, exploring how a baseline that assumes some level of correlation between models can more accurately model the observed distribution of \systemlevel outcomes.
\pl{I think we should provide intuition about what the independent baseline misses...for example, if one individual generates harder examples (due to having a bad microphone/camera), then maybe there's not much you can do; then of course, 'harder'/'easier' is subjective and depends on the model...some accents are 'hard' just because they are less represented rather than being fundamentally harder. Anyway, this would be a good place to discuss these things, because I think this is an important point that is the first thing that will come up, and we shouldn't pretend we have the perfect answer here.}\kc{Added more here and moved the reference to the appendix here.}

Comparing the true observed distribution of \systemlevel outcomes with the \baseline distribution helps illuminate how correlated outcomes are across models.  Below we define $P_{\text{observed}}$ (\autoref{p_observed}) and $P_{\text{\baseline}}$ (\autoref{p_baseline}).
\pl{need some text introducing the equations; otherwise they are just floating and it looks like it's associated with the last sentence}\kc{Introduced equations.}

\begin{align}
    P_{\text{observed}}(t~ \text{failures}) &= \frac{\sum_{i=1}^N \1 \left[t = \sum_{j=1}^k F[i,j] \right]}{N} \label{p_observed} \\  
    P_{\text{\baseline}}(t~ \text{failures}) &= \text{Poisson-Binomial}(\bar{f_1}, \dots, \bar{f_k})[t] \label{p_baseline}
\end{align}


\begin{figure}
    \centering
    \begin{subfigure}[H]{0.5\textwidth}
        \centering
        \includegraphics[width=\textwidth]{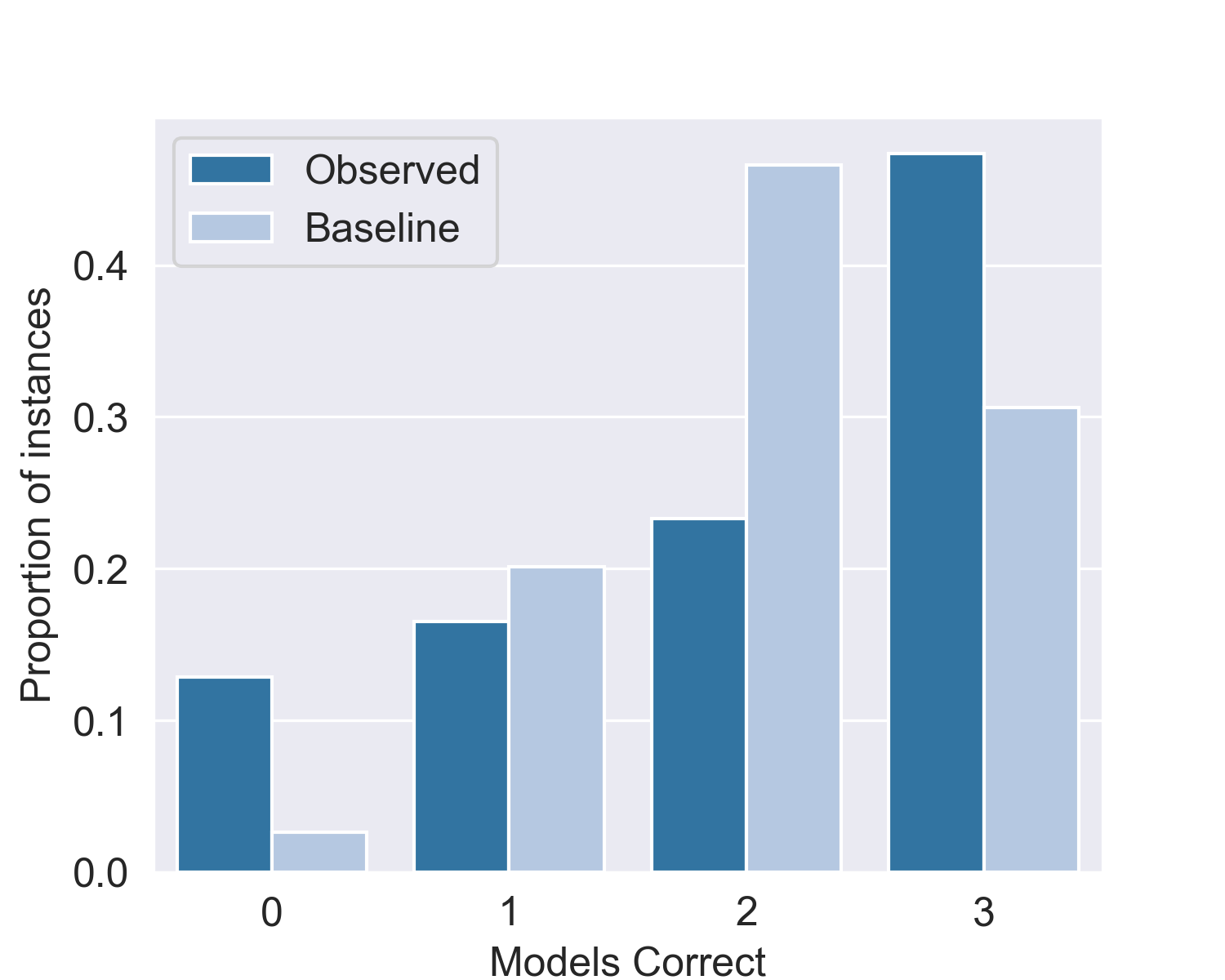}
        \caption{\Systemlevel outcomes for \digit dataset.}
        \label{fig:digit_polarization}
    \end{subfigure}%
    \begin{subfigure}[H]{0.5\textwidth}
        \centering
        \includegraphics[width=\textwidth]{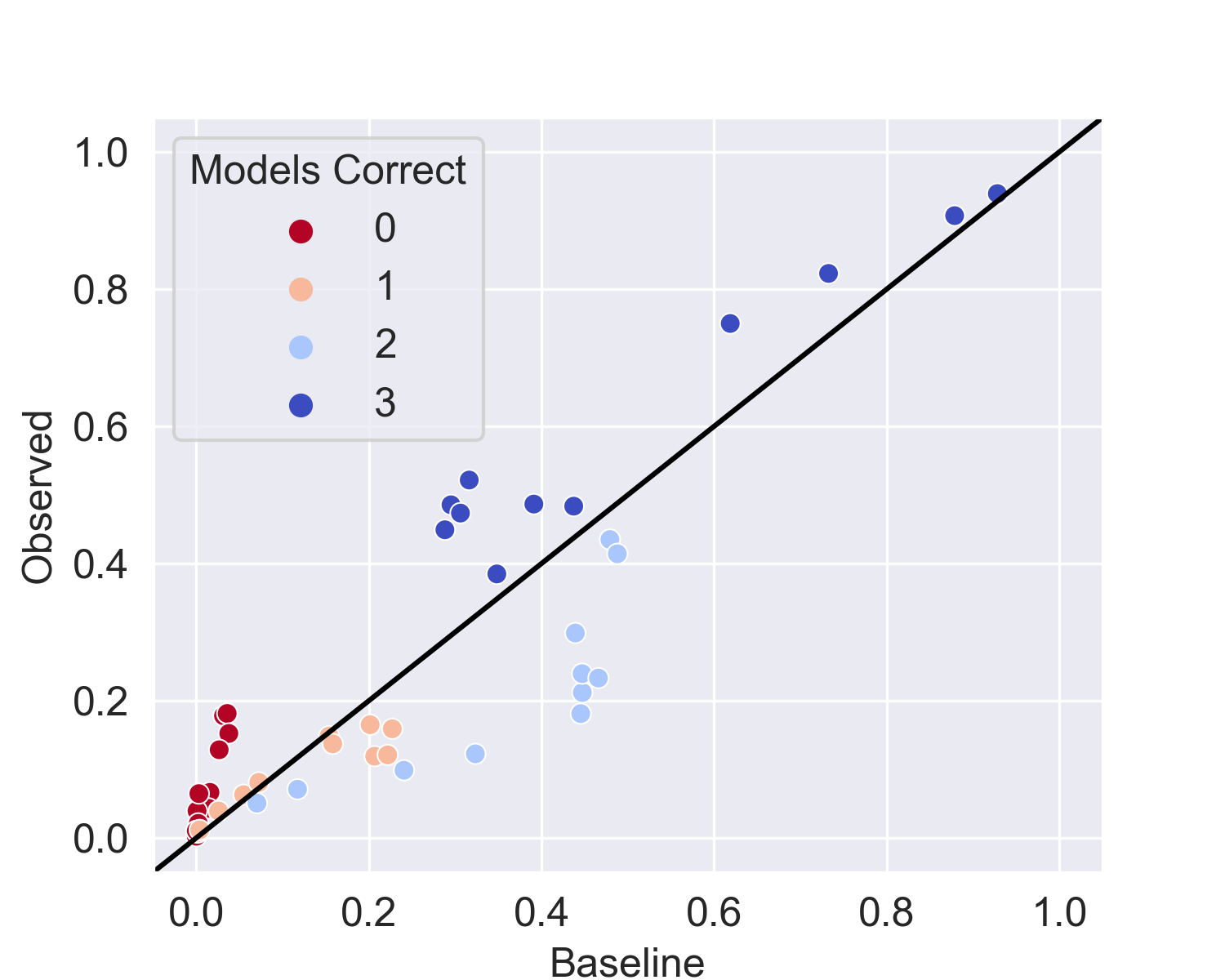}
        \caption{\Systemlevel outcomes for all datasets.}
        \label{fig:polarization_all}
    \end{subfigure}%
    \caption{\textbf{\Profilepolarization.} \Systemlevel analysis surfaces the general trend of \textit{\profilepolarization}: the observed rates that all models succeed/fail consistently exceeds the corresponding \baseline rates.   \autoref{fig:digit_polarization} shows that models in the DIGIT dataset are more likely to all fail or all succeed on an instance than baseline.   \autoref{fig:polarization_all} shows that across all datasets, systemic failure (red dots) and consistent success (blue dots) of all three models on an instance are both more common than baseline, whereas intermediate results are less common than baseline. 
    \pl{need a fuller caption, especially for (b)}\kc{added more here}
    }
\end{figure}

\paragraph{Finding 1: \PolarizedProfiles.} 

\pl{do we need 'polarization' as a word? this just means (positively) 'correlated' in the context of this paper, no?}
\pl{I think polarization has some other connotations which might be confusing; like one might think that it means that some of the models predict one way and some predict another way}\kc{I would be happy to switch to "homogeneous", "extreme and homogeneous", or "correlated" rather than polarized. The current way we use "polarized" is also confusing to me because typically we use "polarized" to describe a whole ecosystem -- e.g. the media has become polarized, political commentary has become polarized. So using "polarized profiles" to describe individual models doesn't feel intuitive.}

In \autoref{fig:digit_polarization}, we compare the observed and \baseline distributions for the spoken command recognition dataset \digit.
We find the observed \systemlevel outcomes are more clearly \textit{\polarized} compared to the baseline distribution: the fraction of instances that receive either extreme outcome (all right or all wrong) exceeds the \baseline rate.
These findings generalize to all the datasets (\autoref{fig:polarization_all}): the observed rate always exceeds the \baseline rate for the \polaroutcomes (above the line $y = x$) and the reverse mostly holds for intermediary outcomes.

\begin{figure}
    \centering
    \includegraphics[width=\textwidth]{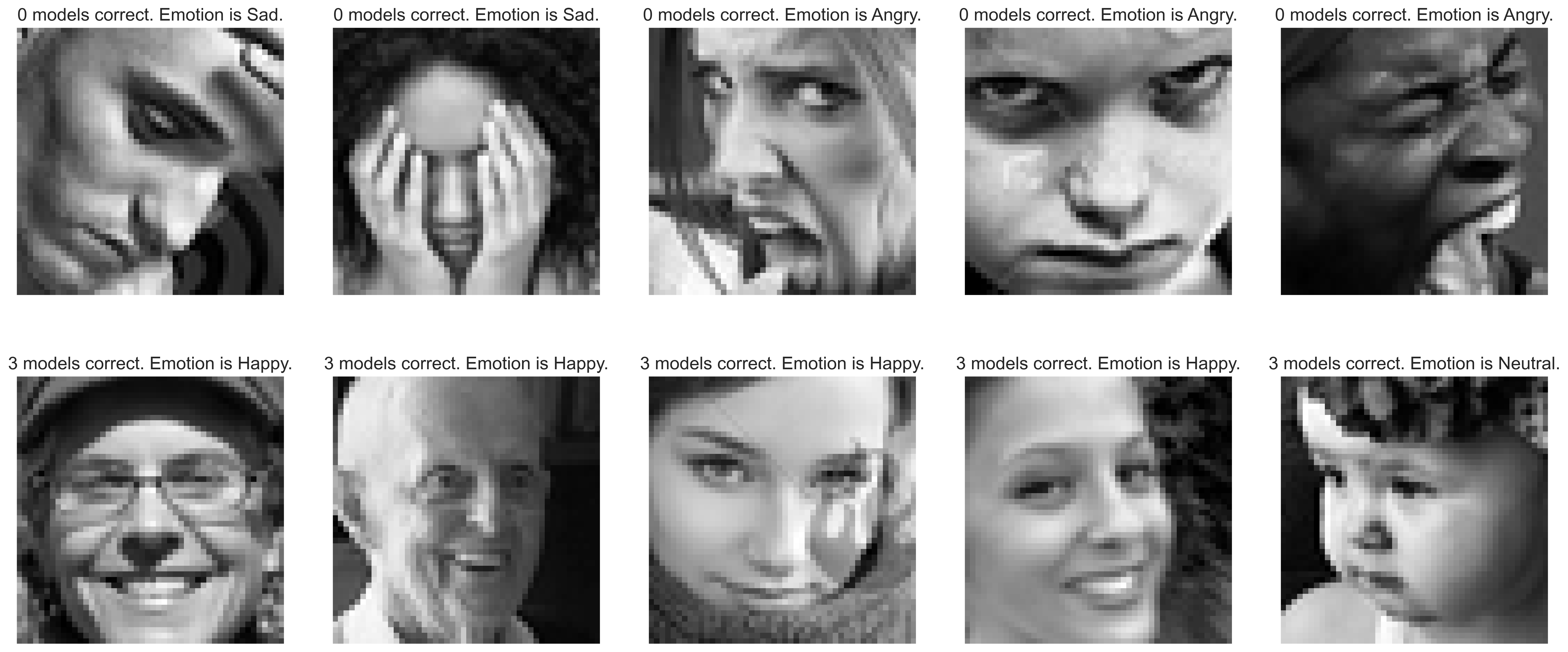}
    \caption{\textbf{Examples of \polaroutcomes} 
    \pl{text at the top is very small - can we make this bigger?}\kc{I think Connor would have to address this.}
    Instances that are sampled uniformly at random from ``0 models correct'' (\textit{top row}) or ''3 models correct'' (\textit{bottom row}) in \ferplus.
    The systemic failures (\textit{top row}) do not appear to be inherently harder for humans to classify; more extensive analysis appears in the supplement.
    }
    \label{fig:FER examples}
\end{figure}

\paragraph{Examples.}
To build further intuition, we present several randomly sampled instances from the \ferplus facial emotion recognition dataset in \autoref{fig:FER examples}.\footnote{We acknowledge the task of facial emotion recognition has been the subject of extensive critique \citep[\eg][]{Barrett2019, LeMau2021}). 
We provide examples for this task due to ease of visualization, but our claims also hold for examples from the text and speech modalities that are provided in the supplement.}
We emphasize that while systemic failures may share structure, we do not believe these instances are \textit{inherently} harder than ones on which models perform well. The authors did not have difficulty labelling these examples, nor would other human labelers. 
We address the question of why systemic failures arise in \autoref{sec:commentary}.
\section{Do Model Improvements Improve Systemic Failures?}
\label{sec:hapi-improvements}

The performance of a deployed machine learning system changes over time.
Developers serve new versions to improve performance \citep{chen2022how, chen2022hapi}, the test distribution shifts over time \citep{koh2021wilds}, and the users (sometimes strategically) change their behavior \citep{2020manipulationproof}.
In spite of this reality, most analyses of the societal impact of machine learning only consider static models.

\Systemlevel analysis provides a new means for understanding how models change and how those changes impact people.
When models change, what are the broader consequences across their model ecosystem?
Do single-model improvements on average improve \systemlevel outcomes by reducing \systemic failures?
And to what extent are the individuals for whom the model improves the same individuals were previously systemically failed?





\paragraph{Setup.}
\citet{chen2022hapi} evaluated the performance of the commercial APIs on the same eleven evaluation datasets each year in 2020--2022.
Of all year-over-year comparisons, we restrict our attention to cases where one of the three APIs for a given task improves by at least 0.5\% accuracy.\footnote{The supplement \autoref{threshold} contains an analysis that confirms our findings are robust to alternate thresholds.} 
Let \himproved denote the model that improved. 
We identify the instances that \himproved initially misclassified in the first year as \textit{potential improvements} and the subset of these instances that \himproved correctly classified in the second year as \textit{improvements}.
Considering the initial distribution of failures for \himproved, we can ask where does the \himproved improve?
We answer this by comparing the distribution of \outcomeprofiles for the \textit{other} models (besides \himproved) between the potential improvement and improvement sets. 

\begin{figure}
    \begin{subfigure}[t][][t]{.5\textwidth}
        \centering
        \includegraphics[width=\textwidth]{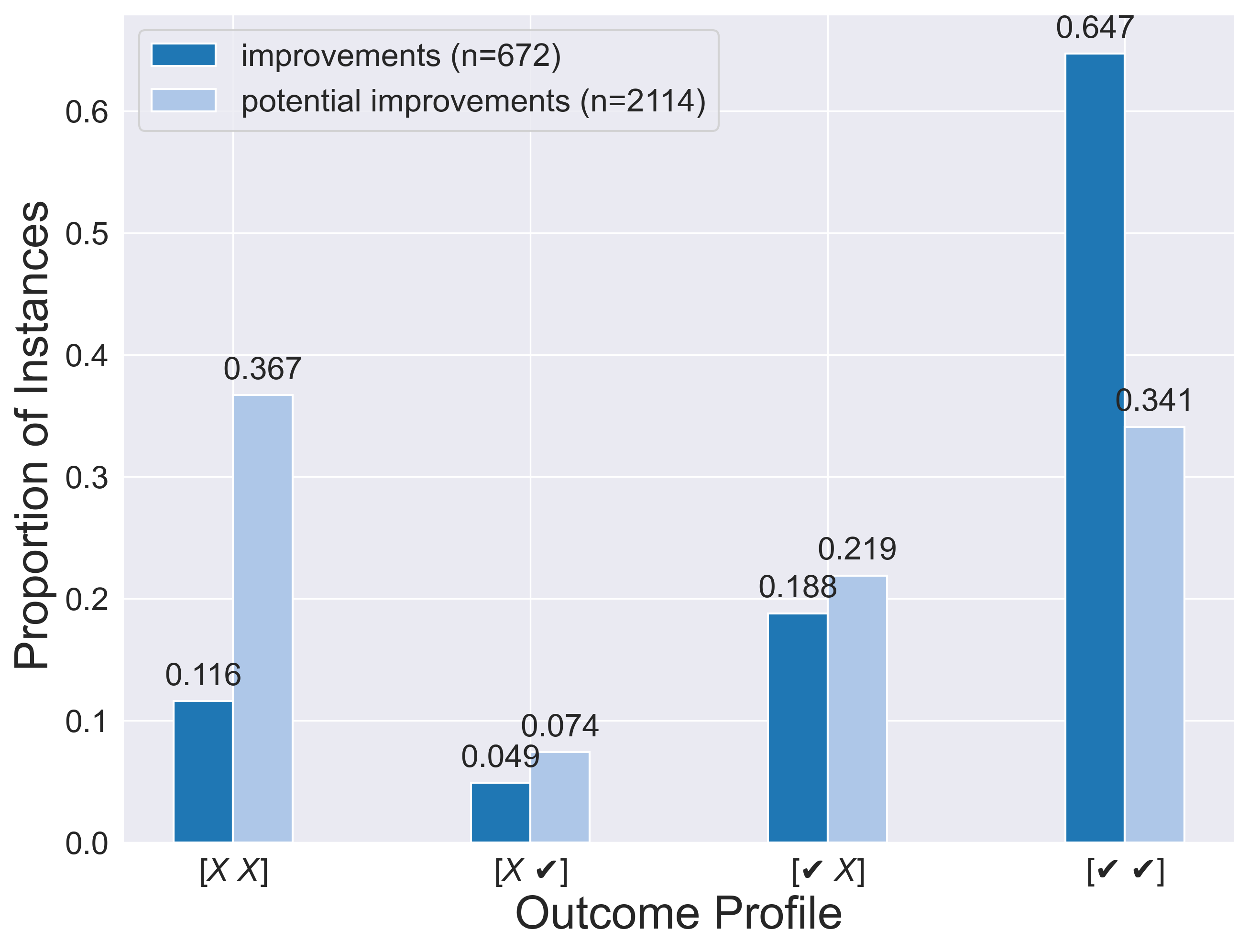}
        \caption{\Outcomeprofile distribution for (Baidu, Google) when Amazon improves on \waimai from 2020 to 2021.}
        \label{fig:WAIMAI improvements}
    \end{subfigure}
    \hspace{1em}
    \begin{subfigure}[t]{.5\textwidth}
        \centering
        \includegraphics[width=\textwidth]{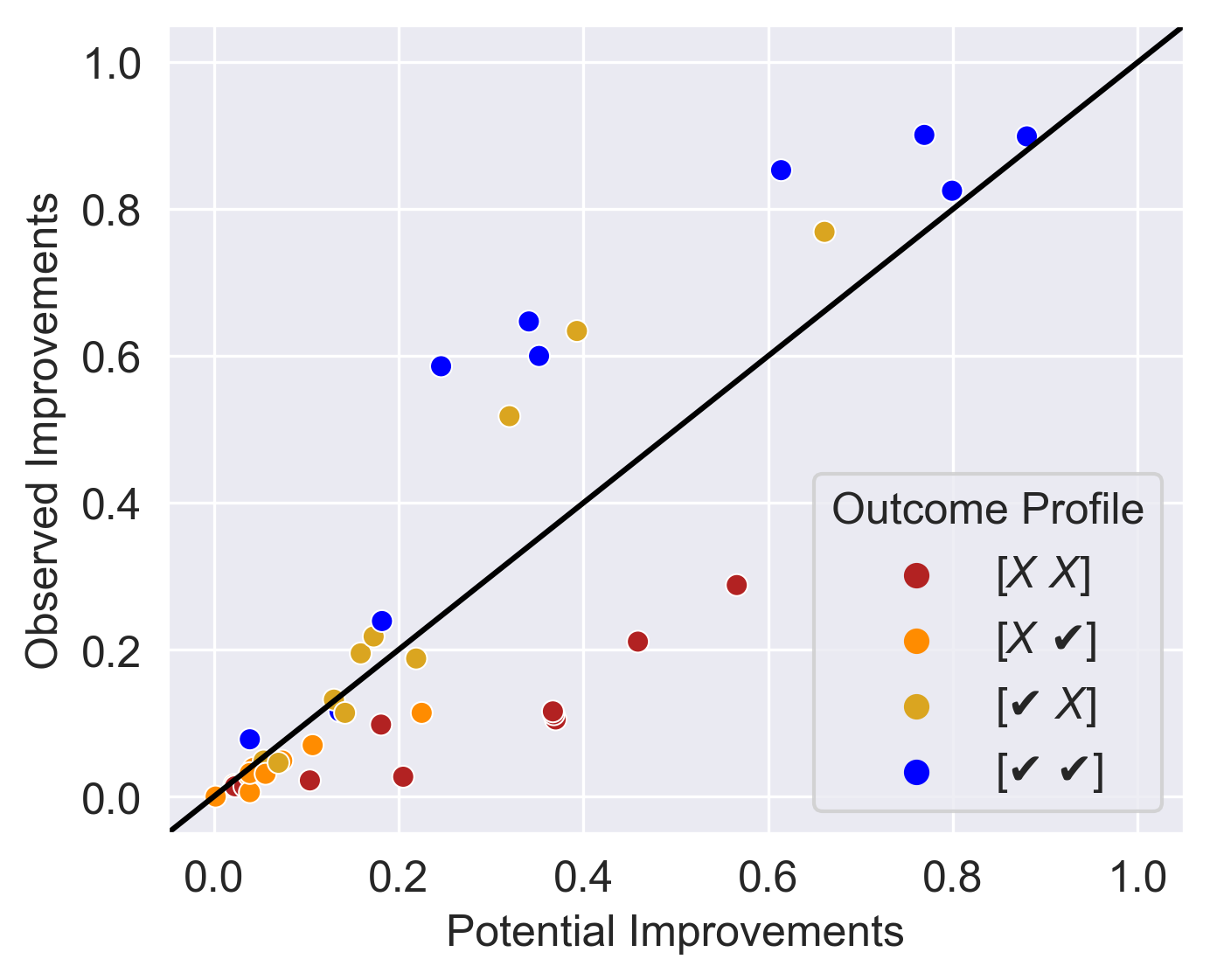}
        \caption{
        The distribution of \outcomeprofiles for all year-over-year model improvements across all datasets.}
        \label{fig:all datasets improvements}
    \end{subfigure}
    \caption{
    \textbf{Model improvement is not concentrated on systemic failures.}
    When a model improves, we compare the distribution of \outcomeprofiles of the other two models on its initial failures (\textit{potential improvements}) to the distribution on the instances it improved on (\textit{observed improvements}). 
    Across all improvements, including Amazon's improvement on \waimai (\textit{left}), there is a clear over-improvement on [\checkmark, \checkmark] (above $y = x$ on \textit{right}) and under-improvement on [X, X] (below the identity line on \textit{right}).
    \pl{I wouldn't say $y=x$ because those symbols are used elsewhere, say black / identity line?}\kc{addressed}
    }
    \label{fig:improvements}
\end{figure}

\paragraph{Finding 2: Model improvements make little progress on systemic failures.}
As a case study, we consider Amazon's improvement on the \waimai dataset from 2020 to 2021.  
In \autoref{fig:WAIMAI improvements}, from left to right, [X, X] indicates the other APIs (Baidu, Google) both fail, [X, \checkmark] and  [\checkmark, X] indicate exactly one of the other APIs succeed, and [\checkmark, \checkmark] indicates both succeed. 
The majority (64.7\%) of Amazon's improvement is on instances already classified correctly by the other two APIs, far exceeding the fraction of potential improvements that were classified correctly by Baidu and Google (34.1\%) in 2020. 
In contrast, for systemic failures, the improvement of 11.6\% falls far short of the potential improvement of 36.7\%. 
In fact, since models can also fail on instances they previously classified correctly, the model's improvement on systemic failures is even worse in terms of net improvement.
 Amazon's improvement amounts to \textit{no net reduction of systemic failures}: the model improves on 78 systemic failures but also regresses on 78 instances that become new systemic failures, amounting to no net improvement.  The Baidu and Google APIs similarly show little improvement on systemic failures even as models improve. 
\pl{it should be the same for Baidu and Google too, right? just don't want to single out Amazon;
you kind of say this in the next paragraph, but I'd do it more directly here}\kc{added a clause and sentences to address this}

This finding is not unique to Amazon's improvement on the \waimai dataset: in the 11 datasets we study, we observe the same pattern of homogeneous improvements from 2020-2022. 
In \autoref{fig:all datasets improvements}, we compare the observed improvement distribution\footnote{The supplement \autoref{netimprovements} contains analysis comparing `net improvements' to `potential improvements' as well; the trends are consistent across both analyses.} ($y$ axis) to the potential improvement distributions ($x$ axis) across all model improvements.
We find a clear pattern: systemic failures (the [X, X] category) are represented less often in the observed improvement set than in the potential improvement set. This finding indicates that when models improve, they \textit{under-improve} on users that are already being failed by other models. 
Instead, model improvements especially concentrate on instances where both other models succeeded already. 

\Systemlevel analysis in the context of model improvements disambiguates two plausible situations that are otherwise conflated: does single-model improvement (i) marginally or (ii) substantively reduce \systemic failures?
We find the reduction of \systemic failures is consistently marginal: in every case, the reduction fails to match the the distribution seen in the previous year (\ie every [X, X] red point is below the line in \autoref{fig:all datasets improvements}). 
\section{\Systemlevel Analysis in Dermatology (DDI)}
\label{sec:dermatology}
Having demonstrated that \systemlevel analysis reveals homogeneous outcomes across machine learning deployments, we apply the \systemlevel methodology to medical imaging.
We consider this setting to be an important use of  \systemlevel analysis because machine learning makes predictions that inform the high-stakes treatment decisions made by dermatologists.

\subsection{Data}
\citet{daneshjou2022disparities} introduced the Diverse Dermatology Images (\ddi) dataset of 656 skin lesion images to evaluate binary classification performance at detecting whether a lesion was malignant or benign.
Images were labelled with the ground-truth based on confirmation from an external medical procedure (biopsy).
In addition, each image is annotated with skintone metadata using the Fitzpatrick scale according to one of three categories: 
Fitzpatrick I \& II (light skin), Fitzpatrick III \& IV (medium skin), and Fitzpatrick V \& VI (dark skin). 
We use the predictions from \citet{daneshjou2022disparities} on \ddi for two prominent ML models (ModelDerm \citep{han2020augment} and DeepDerm \citep{esteva2017dermatologist}) and two board-certified dermatologists.\footnote{\citet{daneshjou2022disparities} also evaluated a third model \citep[HAM10K;][]{tschandl2018ham10000} that almost always predicts the majority class in this class-imbalanced setting. 
We exclude this model since its failures are not interesting, but replicate our analyses in the supplement \autoref{dermatology} to show the findings still hold if it is included.}
We defer further details to  \autoref{dermatology}.

\subsection{Results}

\begin{figure}
    \begin{subfigure}[t]{.5\textwidth}
        \centering
        \includegraphics[width=\textwidth]{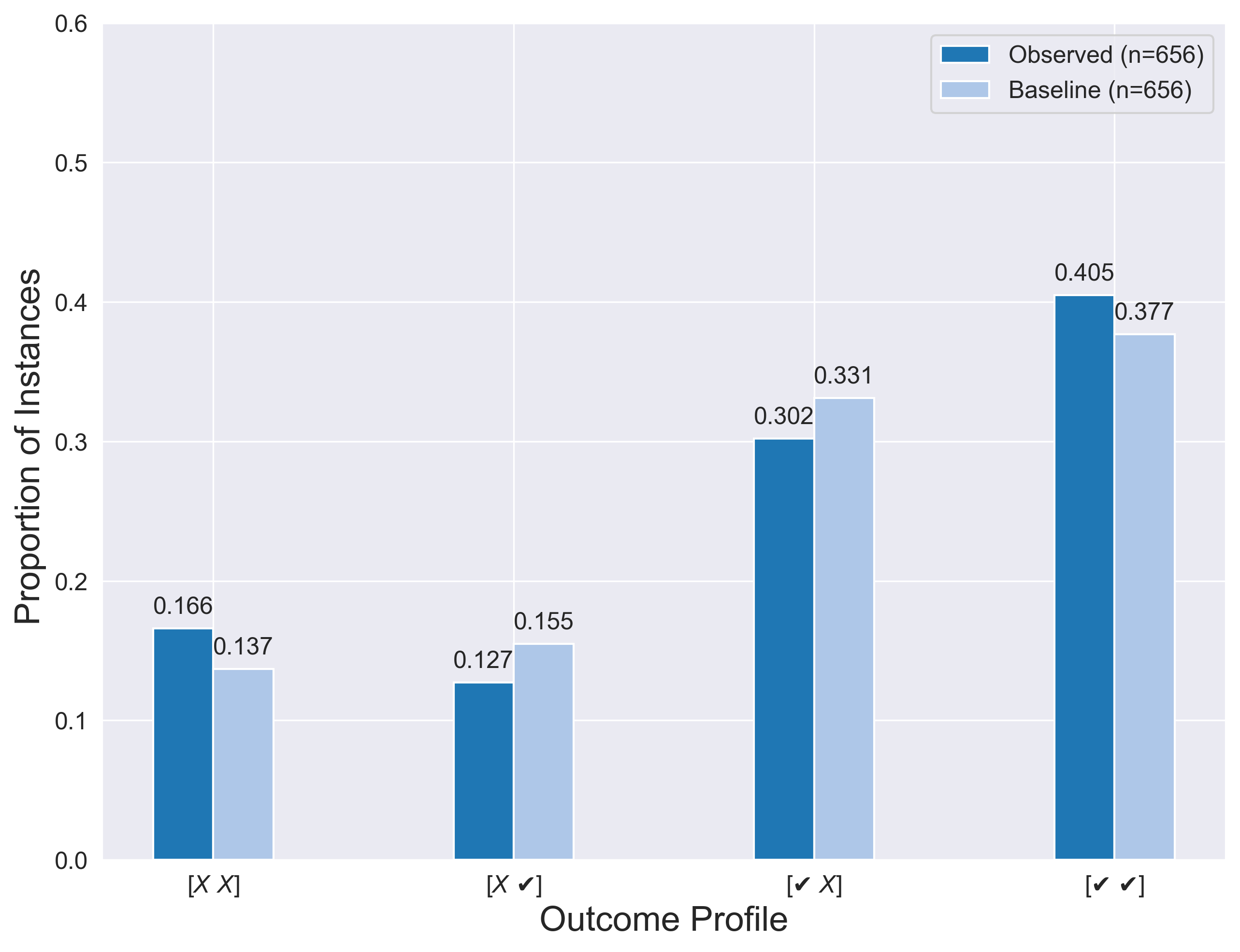}
        \caption{\Outcomeprofiles for models.}
        \label{fig:derm_model_outcomes}
    \end{subfigure}%
    \hspace{1em}
    \begin{subfigure}[t]{.5\textwidth}
        \centering
        \includegraphics[width=\textwidth]{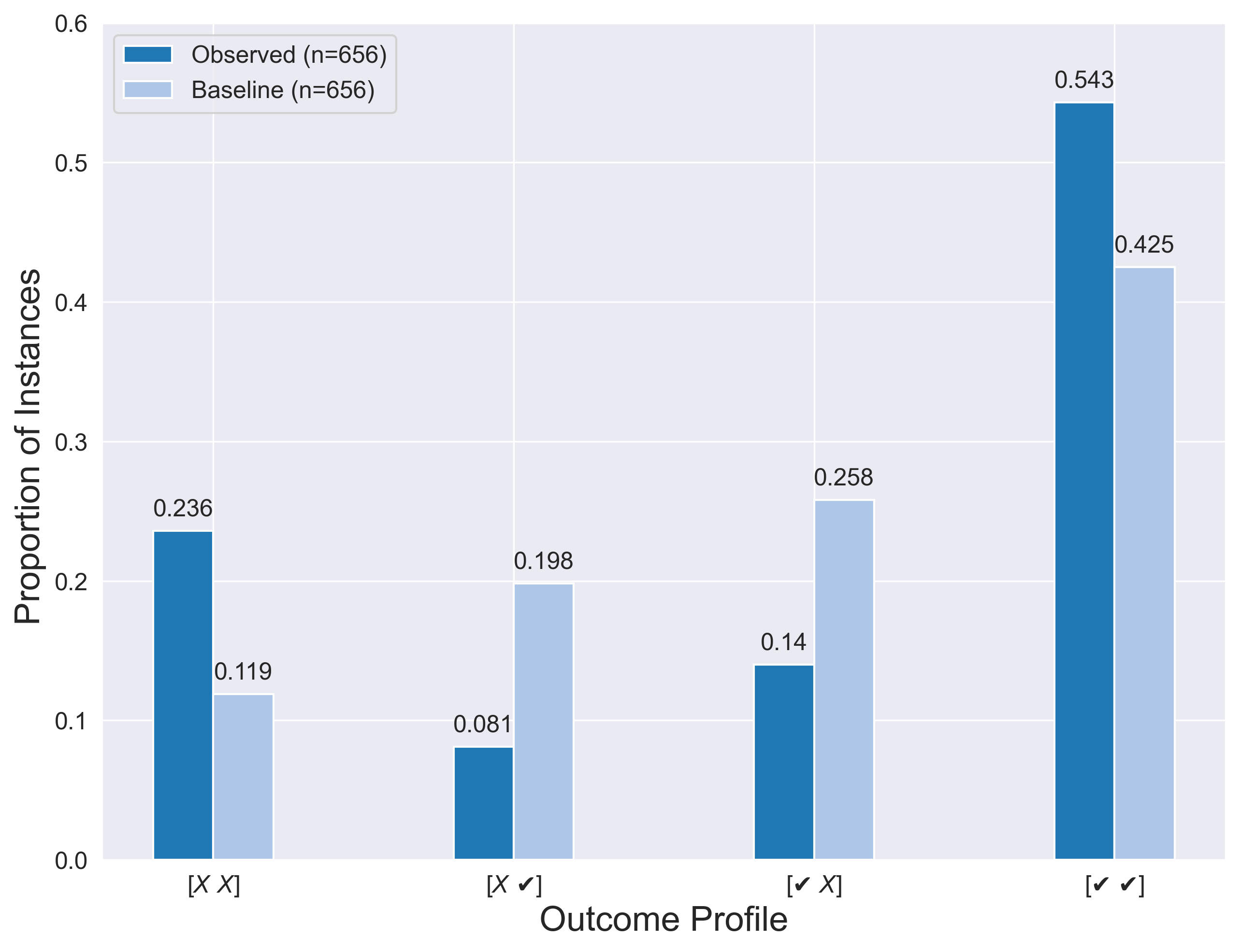}
        \caption{\Outcomeprofiles for humans.}
        \label{fig:derm_human_outcomes}
    \end{subfigure}
    \caption{\textbf{\Profilepolarization for models and humans.} 
    Consistent with \hapi, model predictions (\textit{left}) yield \polarizedprofiles on \ddi. Human predictions (\textit{right}) are even more \polarized than models.}
    \label{fig:human vs model polarization}
    \end{figure}

\paragraph{Finding 3: Both humans and models yield homogeneous outcomes; humans are more homogeneous.} 
We compare \textit{observed} and \textit{\baseline} \systemlevel outcomes on \ddi for models (\autoref{fig:derm_model_outcomes}) and humans (\autoref{fig:derm_human_outcomes}). Consistent with the general trends in \hapi, model predictions yield homogeneous outcomes.
For human predictions from board-certified dermatologists, we also see that outcomes are homogeneous.
However, in comparing the two, we find that humans yield even more homogeneous outcomes.
We take this to be an important reminder: while we predict that models are likely to produce homogeneous outcomes, we should also expect humans to produce homogeneous outcomes, and in some cases more homogeneous outcomes.
\pl{what's the difference between homogenous and polarized outcomes? can we try to standardize to homogenous?}\kc{Yes! Addressed this here and elsewhere.}

\begin{figure}
    \centering
    \begin{subfigure}[c]{.4\textwidth}
        \centering
        \includegraphics[width=\textwidth, height=5cm]{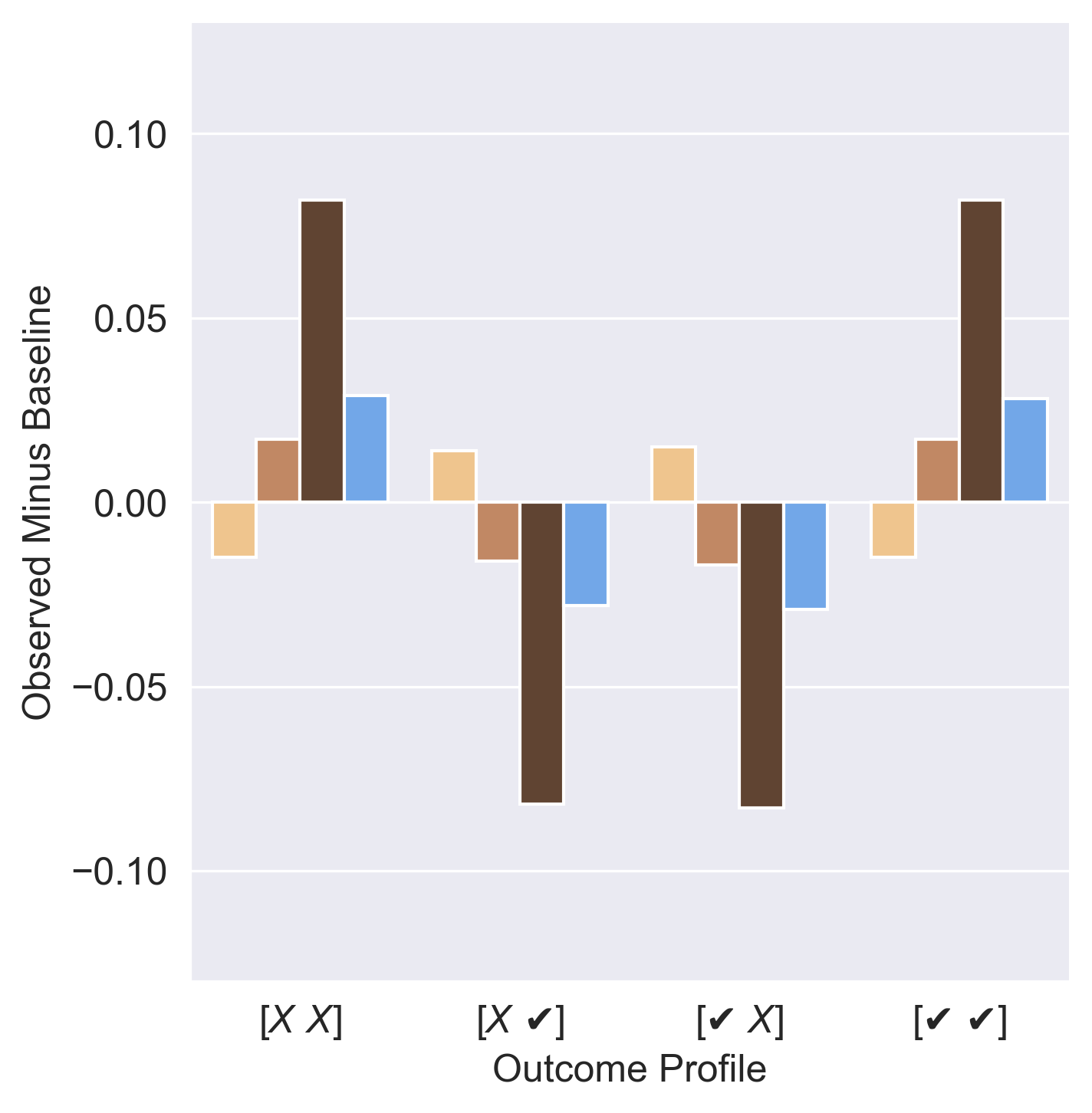}
        \caption{Model \outcomeprofiles by skin tone.}
        \label{fig:diff_by_race_models}
    \end{subfigure}%
    \begin{subfigure}[c]{.4\textwidth}
        \centering
        \includegraphics[width=\textwidth, height=5cm]{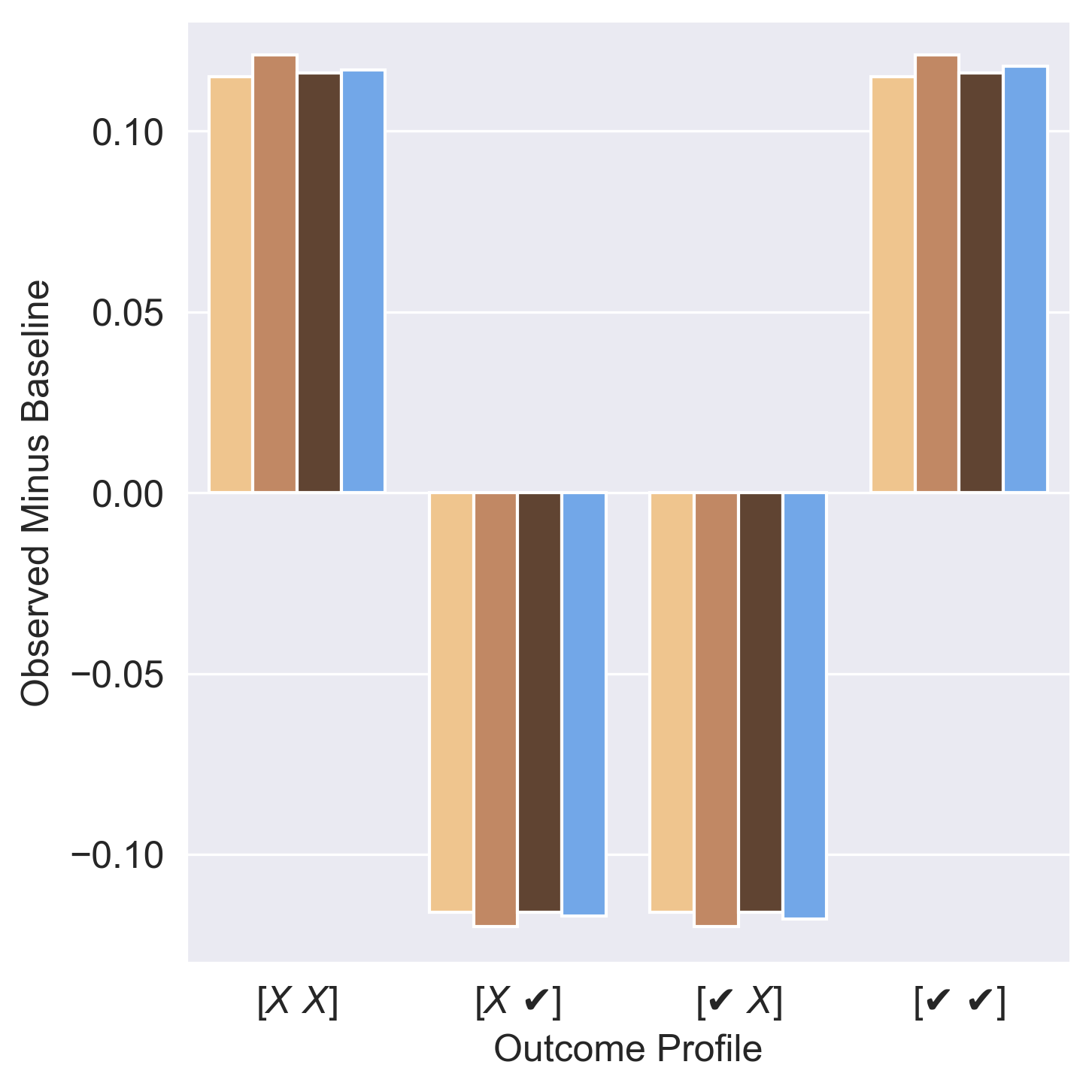}
        \caption{Human \outcomeprofiles by skin tone.}
        \label{fig:diff_by_race_humans}
    \end{subfigure}%
    \begin{subfigure}[c]{.2\textwidth}
        \centering
        \includegraphics[width=\textwidth]{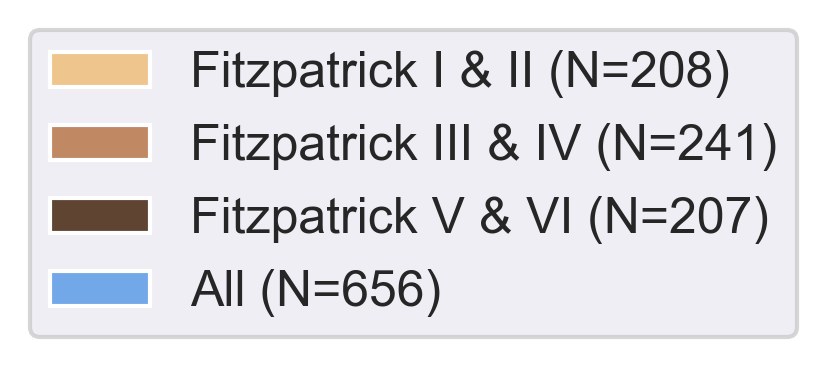}
    \end{subfigure}%
    \caption{
    \textbf{Racial disparities for models but not humans.}
    We stratify \systemlevel analysis in \ddi by the three skin tone categories, plotting the difference between \textit{observed} and \textit{\baseline} rates.
    Models (\textit{left}) show clear racial disparities, exhibiting the most \polarization for the darkest skin tones, whereas humans (\textit{right}) show no significant racial disparity.
    }
    \label{fig:derm_diff_by_race}
\end{figure}

\paragraph{Finding 4: \Systemlevel analysis reveals new racial disparities in models but not humans.} 
Standard analyses of machine learning's impact focus on model performance across groups \citep[\eg][]{buolamwini2018gender, koenecke2020racial}.
In AI for medicine, several works have taken such an approach towards understanding fairness \citep{obermeyer2019dissecting, seyyed2021underdiagnosis, Kim2022, Colwell2022}.  
\citet{daneshjou2022disparities} demonstrate racial disparities for both model and human predictions on \ddi and find that predictive performance is worst for darkest skin tones, aligning with broader trends of racial discrimination in medicine and healthcare \citep{Vyas2020, Williams2015, Fiscella2016}.

\Systemlevel analysis can build on these efforts. We conduct the same analysis from \autoref{fig:human vs model polarization} stratified by skin tone.  In previous experiments, we had observed systemic failures across the whole population.  Here we measure systemic failures for subpopulations grouped by skin tone. 
In \autoref{fig:derm_diff_by_race}, we plot the difference between the observed and \baseline rates on the $y$ axis: the \textbf{All} bars (\textit{blue}) reproduce the \polarization results from \autoref{fig:human vs model polarization}.

\pl{I think this is a really important / interesting finding, so I think it would be helpful to go through it a bit more slowly;
in particular, I'd remind people that we were looking at systemic failures over the entire population.  But what happens now when you look at the systemic failures for each subpopulation defined by skin tone;
Can we also mention the average failure rates across demographic groups? is the average failure disparity lower than systemic failure disparity? 
} \kc{Added two more sentences here -- didn't address the average failure rate yet.}
Strikingly, \systemlevel-analysis surfaces an important contrast between model behavior (\textit{left}) and human behavior (\textit{right}).
Models (\autoref{fig:diff_by_race_models}) are most \polarized for darkest skin tones (Fitzpatrick V \& VI; \textit{dark brown}) and \slightlyantipolarized for the lightest skin tones (Fitzpatrick I \& II; \textit{cream}):  The observed systemic failure rate for the darkest skin tones is 8.2\% higher than the baseline, while for the lightest skin tones it is 1.5\% lower than the baseline. By contrast, humans (\autoref{fig:diff_by_race_humans}) show no significant variation as a function of skin tone.

\Systemlevel analysis therefore identifies a new form of racial disparity not previously documented. Critically, while prior works show racial disparities for both models and humans, here we find a form of racial disparity that is salient for models but not present for humans. We note that, in absolute terms, \polarization is higher for all racial groups in human predictions than model predictions, though human predictions don't display significant differences in \polarization across racial group. This demonstrates that \systemlevel analysis can reveal new dimensions of fairness, allowing stakeholders to identify the metrics that are most relevant in their context, be that model error or systemic failure rate. 
Our tool helps researchers and stakeholders evaluate the holistic ecosystem of algorithmic -- and human -- judgements that ultimately shapes outcomes for those subject to algorithmic judgement.


\section{Commentary}
\label{sec:commentary}

While \systemlevel analysis reveals new dimensions of machine learning's societal impact, it also opens the door for further questions.
We prioritize two here, deferring further discussion and a longer related work section to the supplement.
How can we \textit{explain} the pervasive homogeneous outcomes we have observed in deployed machine learning?
And what are the \textit{implications} of this work for both researchers and policymakers?

\subsection{Explanations for Homogeneous Outcomes} 
Our findings provide evidence across several settings for homogeneous outcomes in deployed machine learning (Finding 1; \autoref{sec:hapi-polarization}) that are mostly unabated by model improvement (Finding 2; \autoref{sec:hapi-improvements}).

\paragraph{Data-centric explanations.} 
We posit that ``example difficulty'' may give rise to homogeneous outcomes and provide three analyses that instantiate variants of this hypothesis.

First, \textit{human ambiguity} on the ground-truth 
may predict homogeneous outcomes.
To test this, we make use of the ten human annotations per example in the \ferplus dataset within \hapi. We find that the systemic failure rate is monotonically correlated with annotator disagreement with the majority label.  This suggests that some systemic failures are correlated with the ambiguity or ``hardness'' of the image.  However, we find that even some of the least ambiguous images, namely images on which all or most annotators agree, have systemic failures.  \pl{why is this the hard case? are you implying that homogenization should be highest for ambiguously labeled examples? connect the dots a bit more}\kc{This wording was incorrect initially; changed to 'least' and added a sentence}.  This indicates that human ambiguity is only partially explanatory of \systemlevel model outcomes.  We explore this further in \autoref{sec:annotator-disagreement}.

Second, \textit{human error} 
may predict homogeneous outcomes.
To test this, we compare human predictions on \ddi with the ground truth biopsy results. We stratify \systemlevel analysis by dermatologist performance, comparing (i) instances both dermatologists get right, (ii) instances they both get wrong, and (iii) instances where they disagree and exactly one is right. 
We find that when both dermatologists fail, there continues to be outcome homogenization. 
However, when both dermatologists succeed, there is no \polarization and the observed rates almost exactly match the \baseline rates for every image.
This suggests human error is also partially predictive of \systemlevel model outcomes. We explore this further in \autoref{sec:human-accuracy}.

Finally, more \textit{expressive theoretical models} can better capture the observed trends than our simple full instance-level independence model.
We introduce a two-parameter model.
$\alpha$ fraction of instances are categorized as difficult and the remaining $1 - \alpha$ are easy.
A model's failure rate $\bar{f_j}$ over all examples scales to $(1 + \Delta)\bar{f_j}$ on hard examples and $\left(1 - \frac{\alpha \Delta}{1 - \alpha}\right)\bar{f_j}$ on easy examples. 
Partitioning examples in this way, while continuing to assume instance-level independence, inherently homogenizes: models are more likely to systemically succeed on easy instances and systemically fail on hard instances. 
To best fit the \hapi data, the resulting average $\alpha$ ($0.2-0.3$) and $\Delta$ ($1-4$, meaning these examples are $2-5\times$ harder) values are quite extreme.
In other words, for this theoretical model to fit the data, a significant fraction ($\approx 25\%$) would need to be considerably harder ($\approx 3.5\times$) than the overall performance. We explore this further in \autoref{sec:expressive-model}.

These analyses contribute to an overarching hypothesis that example \textit{difficulty} partially explains homogeneous outcomes.
While we discuss the construct of difficulty in the supplement, we draw attention to two points.
First, difficulty is largely in the eye of the beholder: what humans or models perceive as difficult can differ \citep[\eg adversarial examples;][]{goodfellow2014adversarial}. 
Thus while example difficulty could be caused by inherent properties of the example (such as noisy speech or corrupted images), it could just as well be due to model properties, such as all the models having made similar architectural assumptions or having parallel limitations in their training data.
Second, whether or not homogeneous outcomes are caused by example difficulty does not change their societal impact.  
The consequences of systemic failure can be material and serious (\eg unemployment).


\paragraph{Model-centric Explanations and Algorithmic Monoculture.} 
An alternative family of explanations put forth in several works is that correlated outcomes occur when different deployed machine learning models share training data, model architectures, model components, learning objectives or broader methodologies \citep{ajunwa2019paradox, Engler2021, bommasani2021, creel2022, fishman2022should, bommasani2022picking, wang2023overcoming}.
Such \textit{algorithmic monoculture} \citep{kleinberg2021monoculture, bommasani2022picking} in fact appears to be increasingly common as many deployed systems, including for the tasks in \hapi, are derived from the same few foundation models \citep{bommasani2023ecosystem, madry2023supplychain}.
Unfortunately, we are unable to test these hypotheses because we know very little about these deployed commercial systems, but we expect they form part of the explanation and encourage future work in this direction.


\paragraph{Implications for Researchers.}
Our paper shows that \systemlevel research can reveal previously-invisible social impacts of machine learning.
We believe this methodology concretizes the impact of decision-making in real contexts in which individuals would typically be affected by decisions from many actors (\eg job applications, medical treatment, loan applications, or rent pricing).
As we demonstrate in \ddi, our methodology applies equally to human-only, machine-only, and more complex intertwined decision-making. We anticipate understanding of homogeneous outcomes arising from any or all of these sources will be valuable.


\paragraph{Implications for Policymakers.}

Given the pervasive and persisting homogeneous outcomes we document, there may be a need for policy intervention.
In many cases no single decision-maker can observe the decisions made by others in the ecosystem, so individual decision-makers (such as companies) may not know that systemic failures exist.  In addition, systemic failures are not currently the responsibility of any single decision-maker, so no decision-maker is incentivized to act alone.
\pl{I'd strengthen: even if you can see what other people are doing (and companies do try out their competitor products and do comparisons internally), the problem is that it's not the responsibility of any individual company structurally (yet)}\kc{Addressed.}
Consequently, policy could implement mechanisms to better monitor \systemlevel outcomes and incentivize \systemlevel improvement.
In parallel, regulators should establish mechanisms for recourse or redress for those currently systemically failed by machine learning \citep[see][]{voigt2017gdpr, cen2023right}.


\section{Conclusion}
\label{sec:conclusion}
We introduce \systemlevel analysis as a new methodology for understanding the cumulative societal impact of machine learning on individuals.
Our analysis on \hapi establishes general trends towards homogeneous outcomes that are largely unaddressed even when models improve.
Our analysis on \ddi exposes new forms of racial disparity in medical imaging that arise in model predictions but not in human predictions.
Moving forward, we hope that future research will build on these empirical findings by providing greater theoretical backing, deeper causal explanations, and satisfactory sociotechnical mitigations.
To ensure machine learning advances the public interest, we should use approaches like \systemlevel analysis to holistically characterize its impact.
\clearpage
\begin{ack}
We would like to thank Shibani Santurkar, Mina Lee, Deb Raji, Meena Jagadeesan, Judy Shen, p-lambda, and the Stanford ML group for their feedback on this work.
We would like to thank James Zou and Lingjiao Chen for guidance with using the \hapi dataset.
We would like to thank Roxana Daneshjou for providing the \ddi dataset along with guidance on how to analyze the dataset.
In addition, the authors would like to thank the Stanford Center for Research on Foundation Models (CRFM) and Institute for Human-Centered Artificial Intelligence (HAI) for providing the ideal home for conducting this interdisciplinary research.
RB was supported by the NSF Graduate Research Fellowship Program under grant number DGE-1655618.
This work was supported in part by a Stanford HAI/Microsoft Azure cloud credit grant and in part by the AI2050 program at Schmidt Futures (Grant G-22-63429).
\end{ack}

\bibliographystyle{unsrtnat}
\bibliography{neurips}
\newpage
\appendix
\section{Data-centric Explanations for Homogeneous Outcomes}
Prior work has explored model-centric explanations for homogeneous outcomes \citep{ajunwa2019paradox, Engler2021, bommasani2021, creel2022, fishman2022should, bommasani2022picking, wang2023overcoming}. 
However, data-centric explanations are comparatively less explored. 
We posit that properties of the underlying data could contribute to the homogeneous outcomes that we observe in all of the datasets we examine.
Intuitively, if we believe that some examples are `hard' and others are `easy', then we might expect to see models all fail for the `hard' examples and all succeed for the `easy' examples.   

We test three variants of this hypothesis.
In \autoref{sec:annotator-disagreement}, we examine how the level of annotator disagreement in the ground truth label impacts \systemlevel behavior.
In \autoref{sec:human-accuracy}, we test how the accuracy of human dermatologists in predicting the malignancy of a skin lesion image correlates with \profilepolarization.
Finally, to build on these finer-grained empirical analyses, in \autoref{sec:expressive-model}, we introduce a more express theoretical model.
Under this model, parameterized by two difficulty parameters, we compute a different \baseline rate for \systemlevel outcomes, showing it can better recover the observation distribution. 

\subsection{Annotator disagreement}
\label{sec:annotator-disagreement}
\begin{figure}
    \centering
    \includegraphics[width=\textwidth]{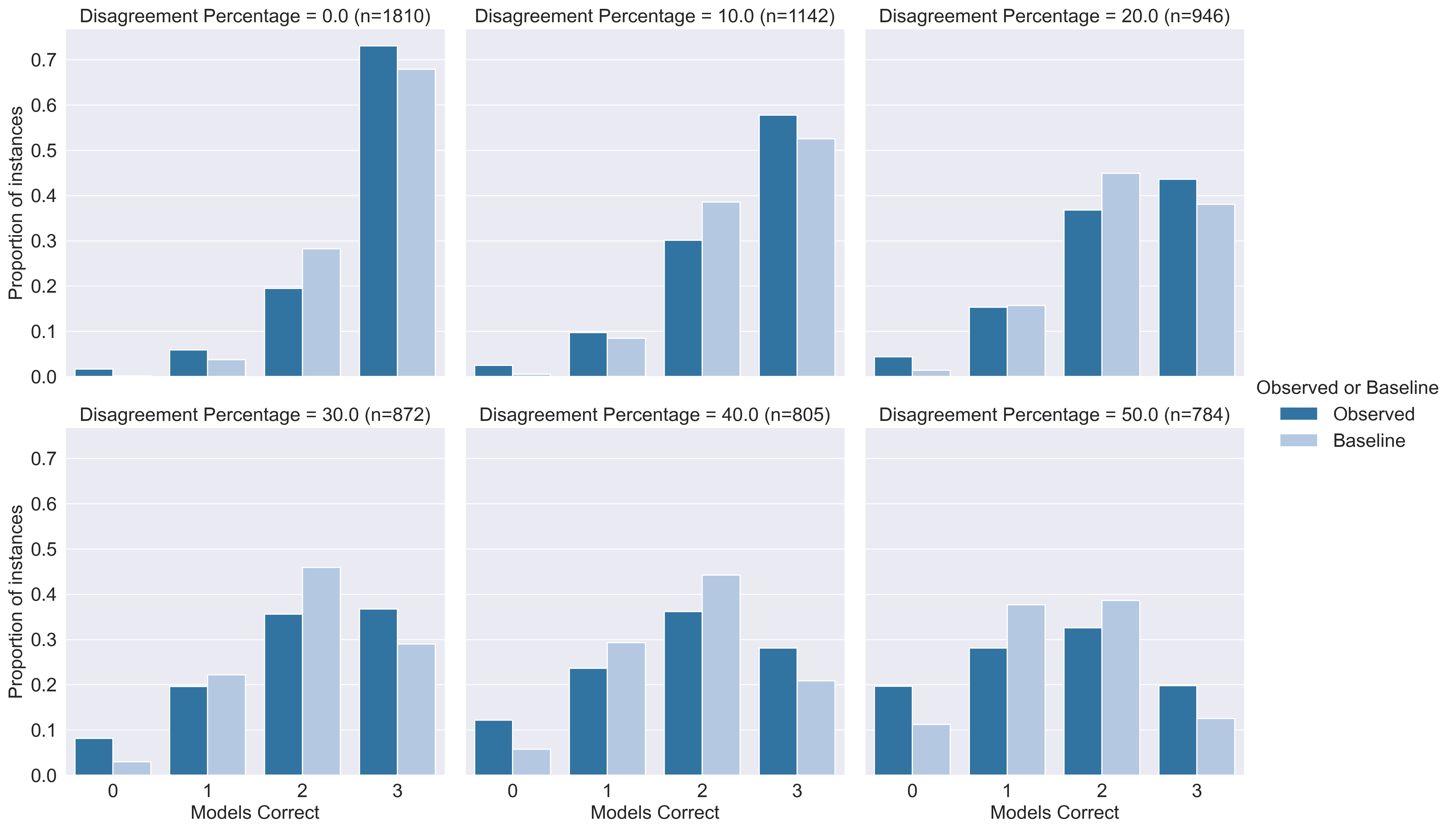}
    \caption{We stratify \systemlevel analysis on \ferplus by instance-level annotator disagreement -- which we take as a proxy for the ambiguity inherent to the input instance -- and plot the difference between \textit{observed} and \textit{\baseline} rates for each instance subset. We observe \profilepolarization for all subsets of data, regardless of the level of annotator disagreement on the instance.
}
    \label{fig:polarization_by_ambiguity}
\end{figure}

To study the effects of annotator disagreement, we make use of the \ferplus dataset.
Each instance of the \ferplus dataset, a facial emotion recognition dataset, contains emotion annotations from 10 human annotators; the emotion label is determined by majority vote \citep{ferplus}. 
Because each instance has been annotated by multiple annotators, we can calculate the annotator disagreement for each instance and use this as a proxy for the `ambiguity' of the instance. 
For example, an instance where all 10 annotators agree that the label is `sad' is less ambiguous than an instance where 6 annotators vote the label should be `fear' and 4 vote that the label should be `surprise'.     

The test set of \ferplus provided in \hapi contains instances with disagreement percentages ranging from 0\% to 50\%. 
We stratify on the disagreement percentage of the instances and compare \textit{\baseline} and \textit{observed} \systemlevel outcomes for each subset of instances.

\paragraph{\PolarizedProfiles manifest regardless of annotator disagreement.} 
In \autoref{fig:polarization_by_ambiguity}, we find that \polarizedprofiles appear for all levels of annotator disagreement.
While more ambiguous examples exhibit higher model error rates and systemic failure rates, the observed rate of \polar outcomes exceeds the \baseline rate of \polar outcomes for all instance subsets. 
This suggests that instance-level ambiguity does not (fully) explain the existence of \polarizedprofiles --- at least in \ferplus. 

\begin{figure}
    \centering
    \includegraphics[width=\textwidth]{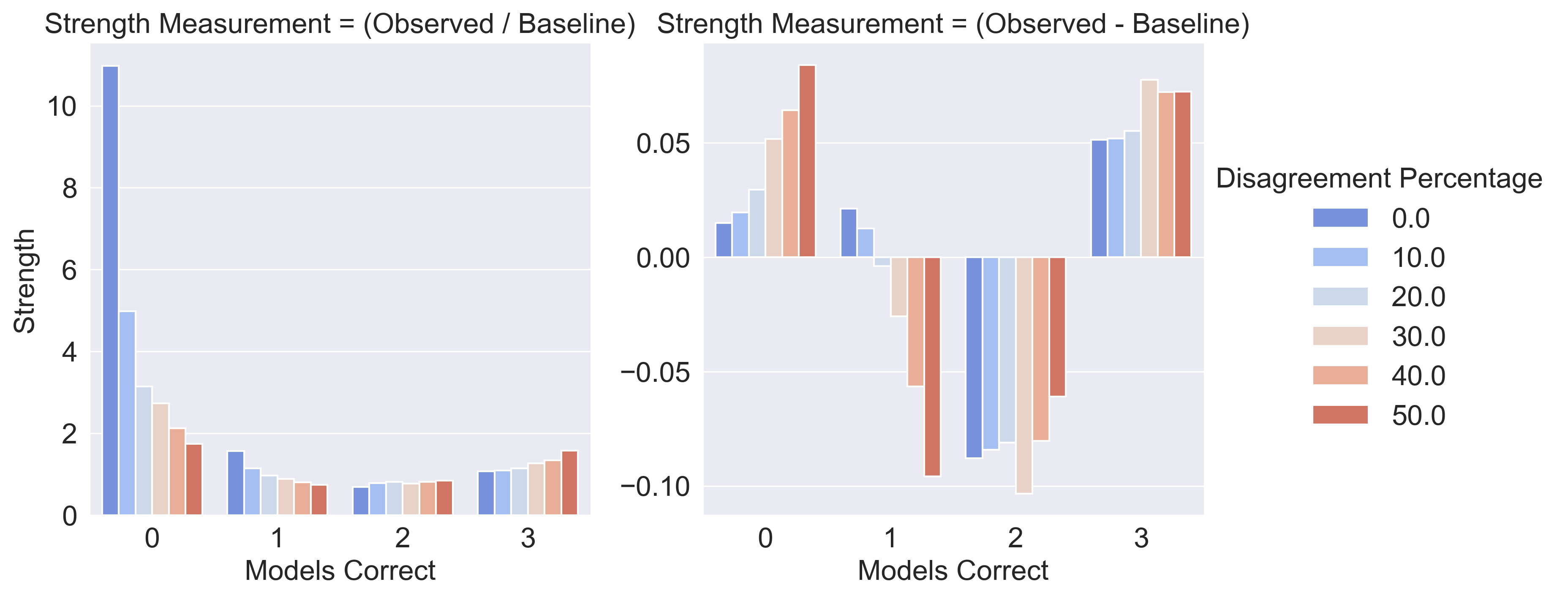}
    \caption{Interpretation of the relationship between instance-level annotator disagreement and \profilepolarization depends on if the strength of the effect is quantified as the \textit{ratio} between observed and \baseline rates (left plot) or as the \textit{difference} between observed and \baseline rates (right plot).
    }
    \label{fig:effect_strength_by_ambiguity}
\end{figure}

\paragraph{The intensity of the effect varies by disagreement level, but the direction of the relationship depends on how strength is quantified.}In light of the observed existence of \profilepolarization across all levels of disagreement, we now examine how the \textit{intensity} of this effect varies with annotator disagreement. The relationship between annotator disagreement and the intensity of \profilepolarization depends on the quantification method used to measure \profilepolarization intensity. When quantifying the intensity as the difference between observed and \baseline rates, the effect becomes more pronounced as disagreement increases. However, when considering the intensity as the ratio between observed and \baseline rates, as in the homogenization metric introduced by \citet{bommasani2022picking}, the effect is most pronounced at lower levels of disagreement.

\subsection{Human accuracy}
\label{sec:human-accuracy}

\begin{figure}
    \begin{subfigure}[t]{.31\textwidth}
        \centering
        \includegraphics[width=\textwidth]{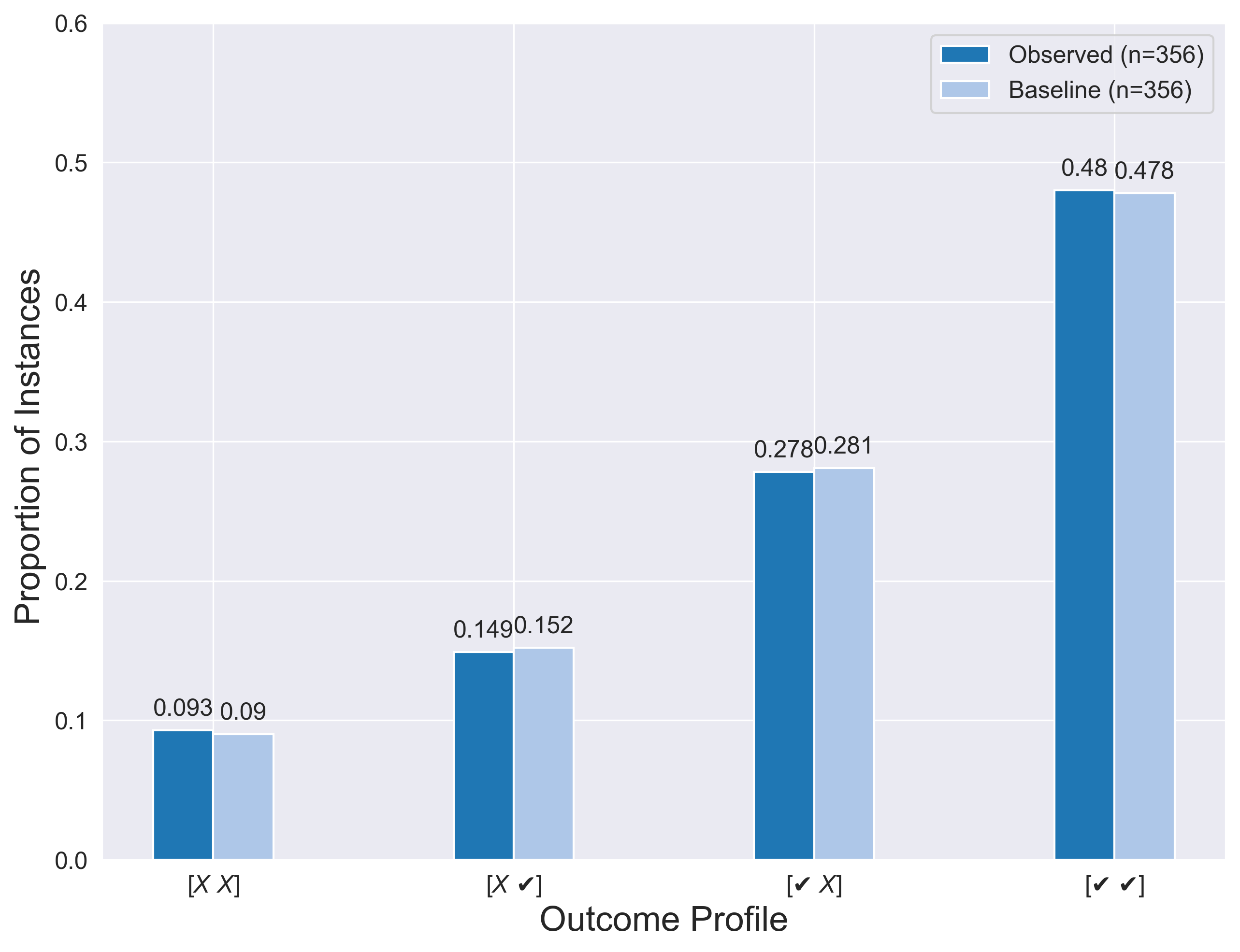}
        \caption{Subset of instances that both dermatologists classify correctly.}
        \label{fig:derm both right}
    \end{subfigure}%
    \hspace{1em}%
    \begin{subfigure}[t]{.31\textwidth}
        \centering
        \includegraphics[width=\textwidth]{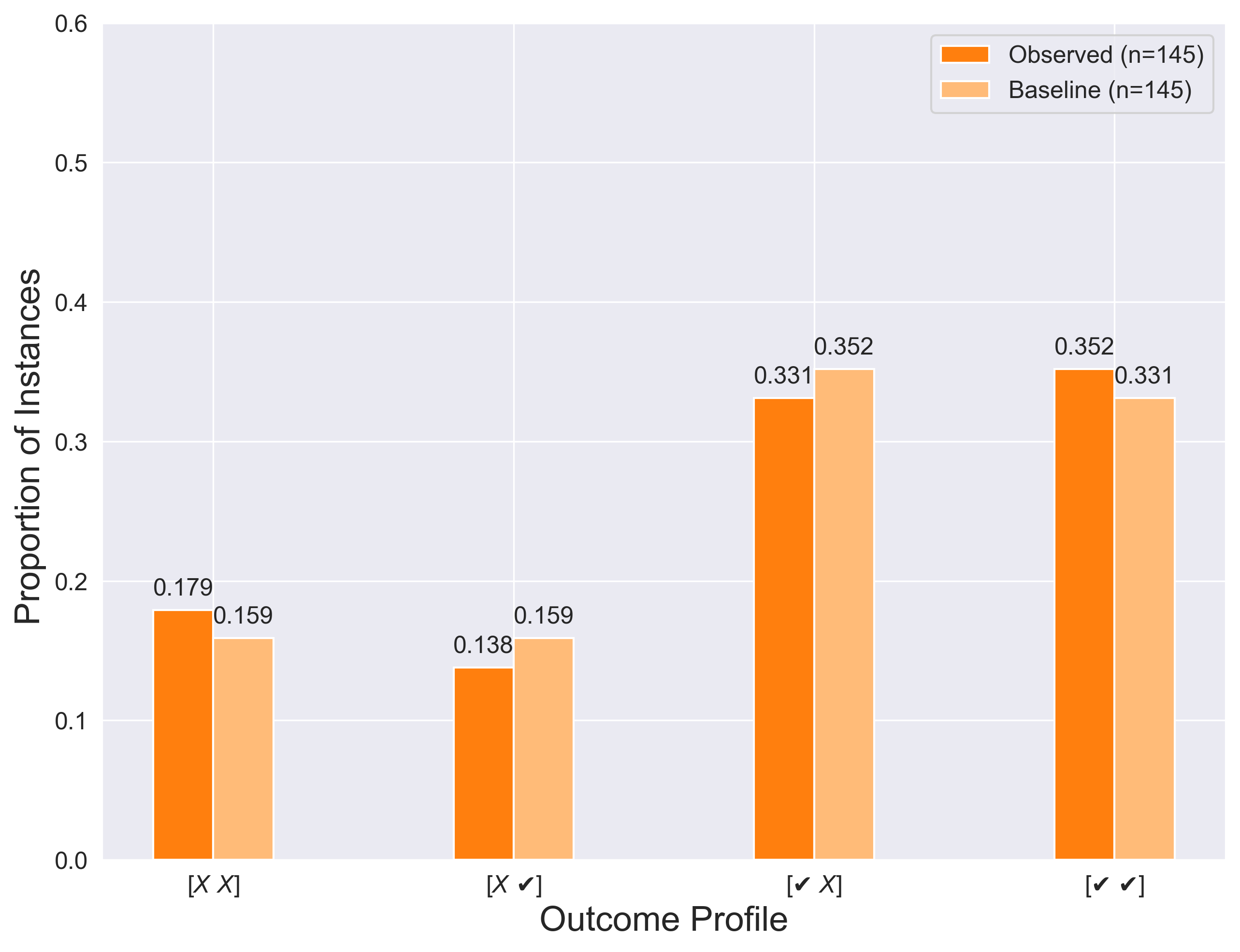}
        \caption{Subset of instances that precisely one dermatologist misclassifies.}
        \label{fig:one derm right}
    \end{subfigure}%
    \hspace{1em}%
    \begin{subfigure}[t]{.31\textwidth}
        \centering
        \includegraphics[width=\textwidth]{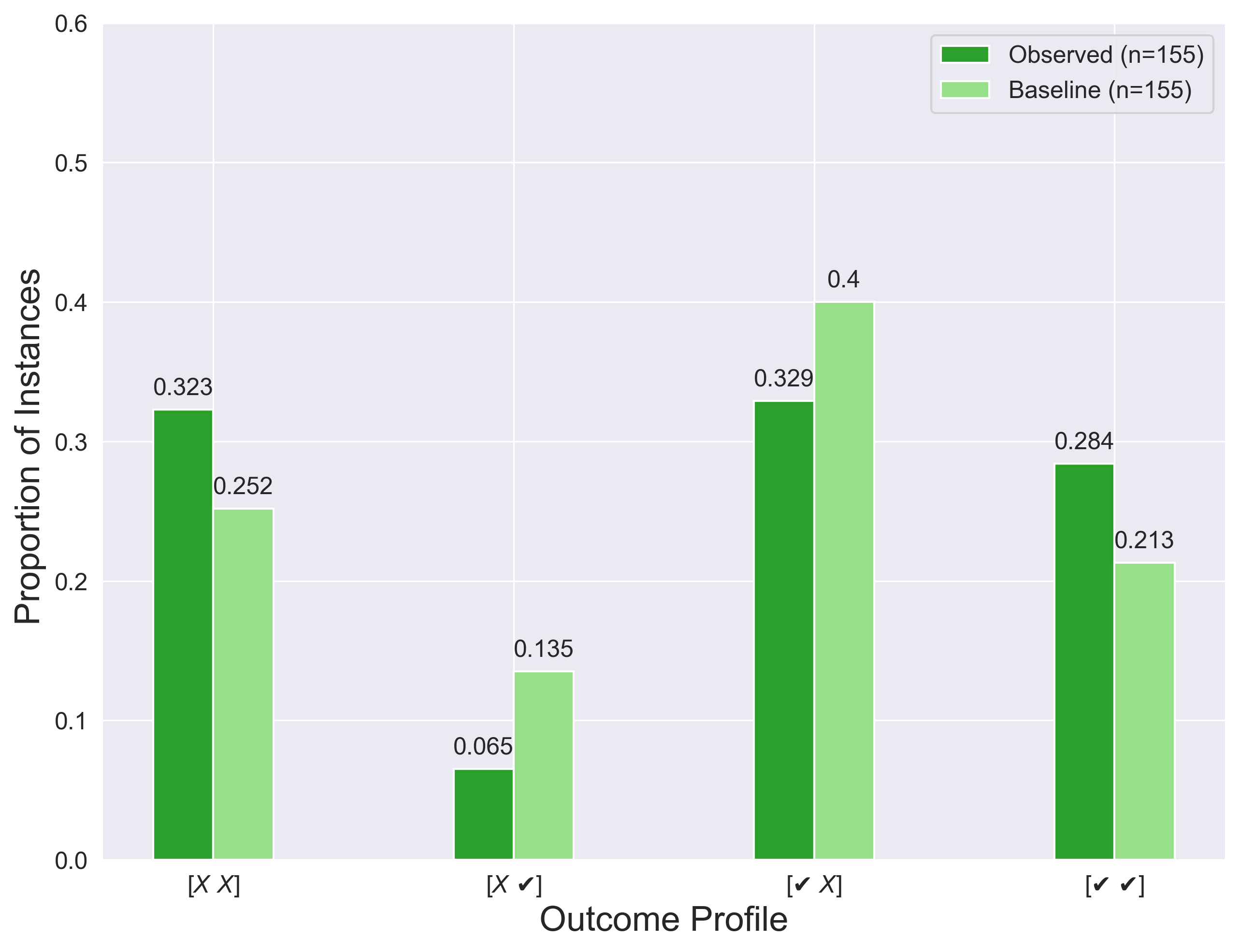}
        \caption{Subset of instances that both dermatologists misclassify.}
        \label{fig:derm both wrong}
    \end{subfigure}%
    \caption{\Profilepolarization exists for the subset of instances where both dermatologists misclassify the image but doesn't exist for the subset where both dermatologists correctly classify the image.}
    \label{fig:derm accuracy polarization}
\end{figure}

Each instance of the \ddi dataset --- a medical imaging skin lesion dataset --- contains predictions from 2 models and 2 humans with the ground truth generated from an external process (in this case, a biopsy of the lesion). To understand how human-perceived difficulty relates to \profilepolarization, we stratify instances on the dermatologist accuracy of that instance (each instance has a dermatologist accuracy of 0\%, 50\%, or 100\%) and examine \profilepolarization for each subset of instances.

\paragraph{Model outcomes are \polarized for instances on which dermatologists fail and \unpolarized for on which dermatologists succeed.} 
In \autoref{fig:derm accuracy polarization}, we find that model outcomes exhibit \profilepolarization for the subset of instances that both human dermatologists fail at to an even greater extent than observed at the population level. 
In contrast, the observed and \baseline distributions match each other (meaning there is no \profilepolarization, \antipolarization or other form of deviation between the distributions) for the subset of instances that both dermatologists get right. 
Note that in the instances that both dermatologists fail at, both models systemically fail more than the \baseline, but both also succeed more than the \baseline. While the correlation between human systemic failures and model systemic failures seems intuitive, it's less clear why models jointly succeed more than the \baseline on human systemic failures. 

Given these surprising and unintuitive findings, we encourage future work to explain what features of these instances lead models to pattern together.
We speculate that these instances lack sufficient informational cues, prompting the models to place excess emphasis on a narrow selection of features, consistent with the literature of machine learning models being susceptible to spurious correlations \citep[\eg][]{sagawa2020investigation}.
At face value, these results would suggest that human-level difficulty can be predictive of outcome \polarization, but we emphasize two caveats:
(i) the \ddi dataset's sample size is limited (155 instances of dermatologists both failing and 356 instances of dermatologists both succeeding) and
(ii) we rely on the judgments of just two human domain experts, meaning the findings may not generalize to larger annotator pools of non-experts (as are common in many NLP or computer vision datasets).
In general, we caution against overgeneralizing this finding. We provide it as an initial foray into understanding example difficulty in a unique setting where we have data that supports an approximation of human-perceived difficulty.

\subsection{More expressive theoretical models}
\label{sec:expressive-model}
 Finally, while we consistently find that observed \systemlevel behaviors yield more homogeneous outcomes than the instance-level independent model predictions would predict, we might intuit there exists instance-level structure that should be encoded into the prior. Therefore, we consider a theoretical framework to encode richer priors on what we might expect of models.
As a simple model, we will assume some instances are universally `hard', meaning all models will perform worse on average across these instances, and others are conversely `easy`, meaning all models will perform better on average across these instances. 

As before, we note that there is not a universal standard according to which examples can be considered `hard.' Some examples are easy for humans but hard for models; other examples are easy for models but challenging for humans; and still other examples are challenging for some models but not others. The three hypotheses we consider in this section explore these observer-relative dimensions of hardness. 

We use two parameters ($\alpha, \Delta$) to parameterize this model, thereby adjusting the \baseline rate that we calculate for \systemlevel outcomes. 
$\alpha$ specifies the composition of `hard' vs `easy' instances in a dataset and $\Delta$ controls how much harder or easier the hard or easy examples, respectively, are expected to be. 
Concretely, $\alpha$ fraction of instances are categorized as difficult and the remaining $1 - \alpha$ are easy. 
A model's failure rate $\bar{f_j}$ over all examples scales to $\bar{f_j}^{\text{hard}} = (1 + \Delta)\bar{f_j}$ on hard examples and $\bar{f_j}^{\text{easy}} = \left(1 - \frac{\alpha \Delta}{1 - \alpha}\right)\bar{f_j}$ on easy examples. 

The distribution of the \textit{\baseline} number of model failures $t \in \{0, \dots, k\}$ follows a weighted sum of two Poisson-Binomial  distributions parameterized by the scaled hard $\bar{f_j}^{\text{hard}}$ and easy $\bar{f_j}^{\text{easy}}$ model error rates. 
\begin{align}
    P_{\text{\baseline}}^{\text{hard}}(t~ \text{failures}) &= (\alpha) \text{Poisson-Binomial}(\bar{f_1}^{\text{hard}}, \dots, \bar{f_k}^{\text{hard}})[t] \\
    P_{\text{\baseline}}^{\text{easy}}(t~ \text{failures}) &= (1 - \alpha) \text{Poisson-Binomial}(\bar{f_1}^{\text{easy}}, \dots, \bar{f_k}^{\text{easy}})[t] \\
    P_{\text{\baseline}}(t~ \text{failures}) &= P_{\text{\baseline}}^{\text{hard}}[t] + P_{\text{\baseline}}^{\text{easy}}[t]
\end{align}

\begin{figure}
    \centering
    \includegraphics[width=\textwidth]{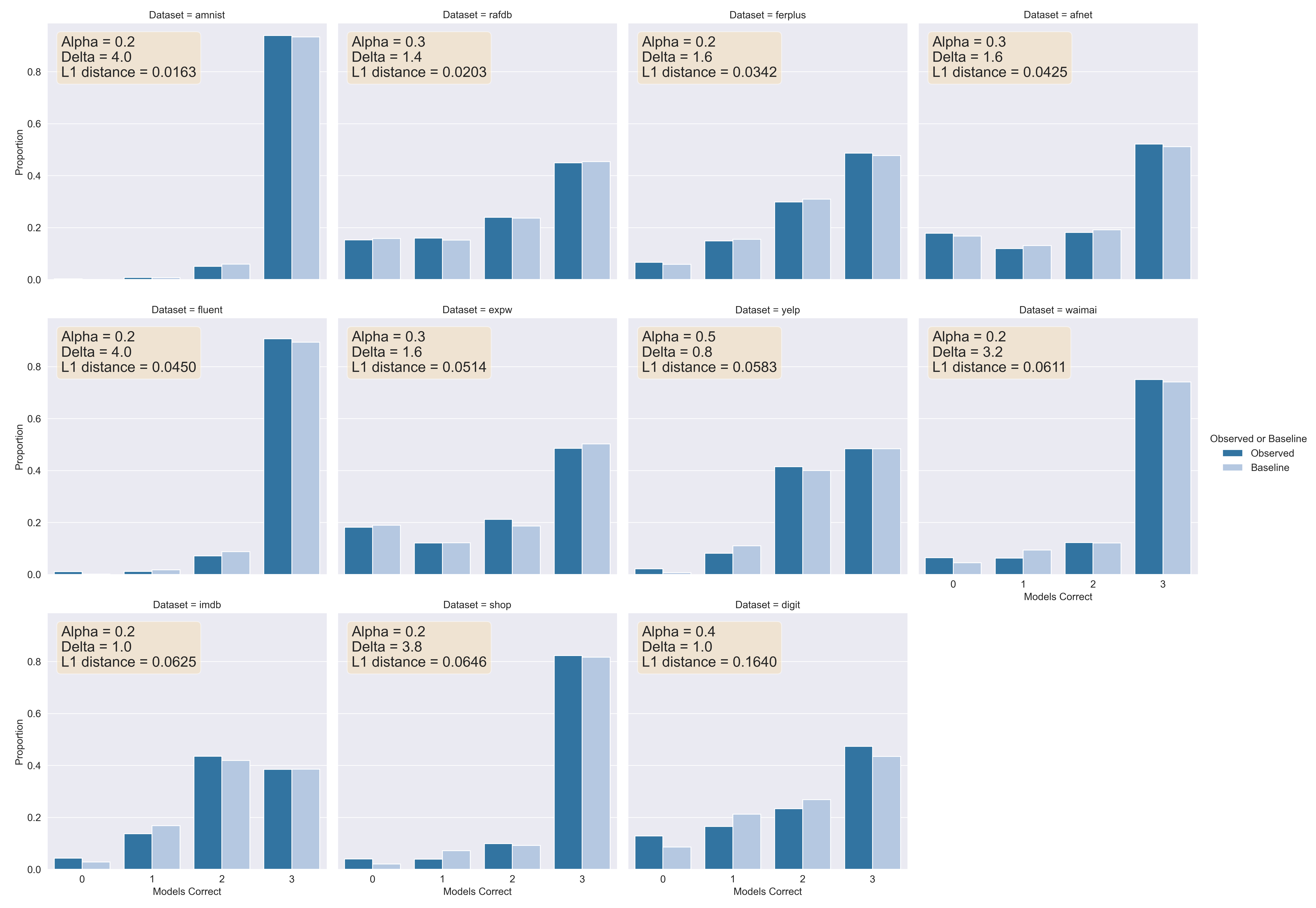}
    \caption{Observed and \baseline distributions of \systemlevel outcomes for the $(\alpha, \Delta)$ combination that yields the lowest L1 distance for each dataset.}
    \label{fig:best_alpha_delta_distributions}
\end{figure}

\paragraph{Identifying $\alpha$ and $\Delta$ values that recover the observed \systemlevel outcomes in \hapi.} We utilize this framework to identify which $(\alpha, \Delta)$ combinations generate \baseline distributions that recover the observed distributions in the \hapi datasets. We perform a grid search for $\alpha \in [0.1, 0.5]$ with a step size of $0.1$ and $\Delta \in [0.2, 5]$ with a step size of $0.2$. Note that certain $(\alpha, \Delta)$ combinations can result in invalid error rates depending on the original error rates of models -- \ie when $\left(1 - \frac{\alpha \Delta}{1 - \alpha}\right)\bar{f_j} < 0$. 

\begin{figure}
    \centering
    \includegraphics[width=\textwidth]{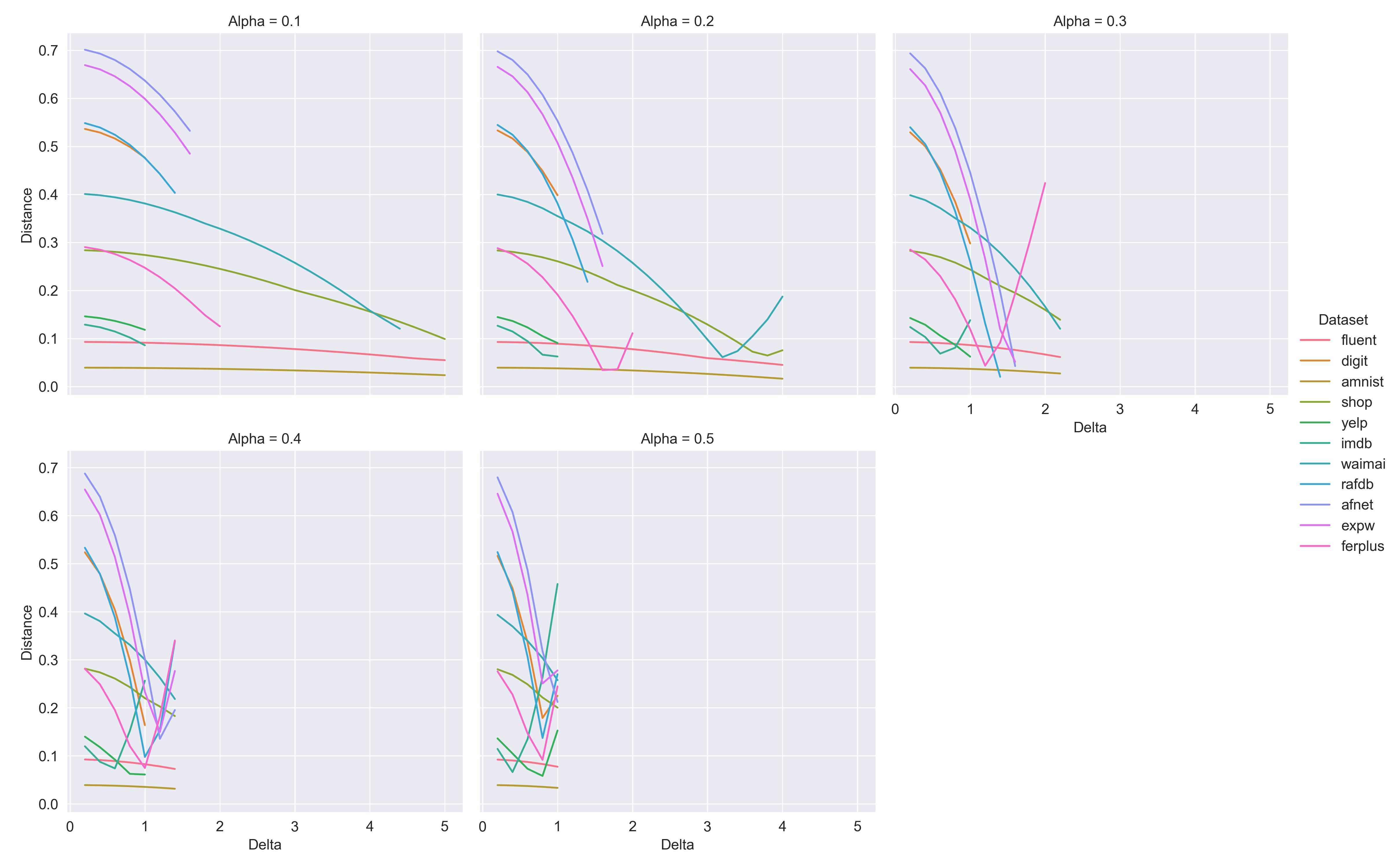}
    \caption{L1 distance between observed and \baseline distributions for all datasets as a function of $\alpha$ and $\Delta$.}
    \label{fig:l1_distances}
\end{figure}

For each dataset, we identify the $(\alpha, \Delta)$ combination that minimizes the L1 distance (equivalently the total variation distance) between the observed and \baseline distributions. 
In \autoref{fig:best_alpha_delta_distributions}, we visualize the observed and \baseline distributions for the distance-minimizing $(\alpha, \Delta)$ pair for each dataset, and in \autoref{fig:l1_distances} we plot the L1 distance for each dataset as a function of $\alpha$ and $\Delta$. 
The majority of the best $\alpha$ values are $0.2$ or $0.3$ while the $\Delta$ values can range from $1$ to $4$ -- indicating that the $\bar{f_j}^{\text{hard}}$  can be up to 5x higher than $\bar{f_j}$. 

\paragraph{High $\Delta$ and low $\alpha$ represents a small group of very difficult instances that all models struggle at.} The combination of high $\Delta$ and low $\alpha$ values performing well suggests that we would need to expect that a relatively small fraction of the dataset contains instances that are significantly harder for all models to perform well at. Note that this framework assumes that all models consider the same instances to be `hard': this agreement is a form of homogenization that could be caused about something inherent to the data or something about how the models are constructed.

\section{\hapi Experiments}
In the main paper, we work extensively with the \hapi dataset of \citet{chen2022hapi}.
While we defer extensive details about the dataset to their work, we include additional relevant details here as well as any relevant decisions we made in using the \hapi dataset.

\subsection{Data}
We work with a subset of \hapi, a dataset introduced by \citet{chen2022hapi} which contains predictions from commercial ML APIs on a variety of standard benchmark datasets from 2020-2022. 
\hapi contains benchmark datasets for three classification tasks and three structured prediction tasks. We only work with the classification tasks -- spoken command recognition (SCR), sentiment analysis (SA), and facial emotion recognition (FER) -- because they contain a single ground truth label where the notion of a `failure' is clear (\ie a misclassification).

The three classifiation tasks span 3 modalities (text, images, speech), and each task is associated with 4 datasets.
However, we exclude one of these datasets: the \command dataset has duplicate example IDs with differing predictions from the same ML API provider in the same time period. 
This makes calculating \systemlevel outcomes impossible. 
After excluding the \command dataset, there are 11 datasets that we conduct \systemlevel analysis on: \rafdb \citep{rafdb}, \afnet \citep{afnet}, \expw \citep{expw}, \ferplus \citep{ferplus}, \fluent \citep{fluent}, \digit \citep{digit}, \amnist \citep{amnist}, \shop, \footnote{\url{https://github.com/SophonPlus/ChineseNlpCorpus/tree/master/datasets/online_shopping_10_cats}} \yelp, \footnote{\url{https://www.kaggle.com/datasets/yelp-dataset/yelp-dataset}}, \imdb \citep{imdb} and \waimai. \footnote{\url{https://github.com/SophonPlus/ChineseNlpCorpus/tree/master/datasets/waimai_10k}}

Each dataset contains predictions from 3 commercial ML APIs in 2020, 2021, and 2022; however the \hapi API did not return predictions from the Face++ model on \afnet in 2022, so we use 2021 predictions for \afnet when conducting experiments that only use predictions from a single year. 

The \hapi dataset is distributed at \url{https://github.com/lchen001/HAPI} under Apache License 2.0.

\subsection{Leader Following Effects in Systemic Failure}
\begin{figure}
    \centering
    \includegraphics[width=\textwidth]{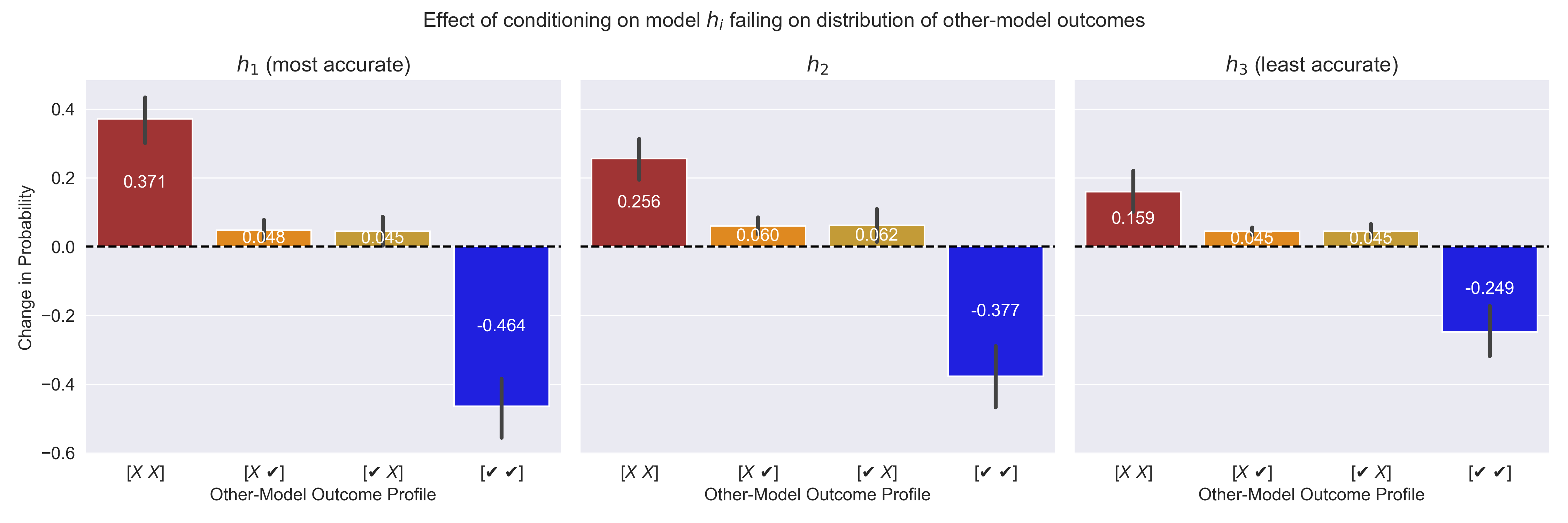}
    \caption{Change in the probability of observing each outcome profile for the other models upon observing $h_i$ fail.}
    \label{fig:leader_following}
\end{figure}
One consequence of \profilepolarization is that it concentrates failures on the same users, so a user who interacts with a model and experiences a failure from that model is now more likely to experience a failure from every other model in the ecosystem. To quantify the strength of this effect, we examine how the probability of a user experiencing each outcome profile changes after that user experiences one failure from a model.

In \autoref{fig:leader_following}, we find that, consistent with the observed \profilepolarization in \hapi, observing a single model failure significantly increases the probability that the user will now experience failures from all other models in the ecosystem. However, we also find that the strength of this effect is strongly graded by the accuracy of the model for which we initially observe a failure. Upon observing the most accurate model in the ecosystem fail for a user, the probability of that user experiencing a systemic failure increases by 37\% -- whereas it only increases by around 16\% when we observe the least accurate model fail.

This result suggests that instances that the most accurate model fails on are likely to be failed by the less accurate models as well. This `leader following' phenomenon has implications for users in a model ecosystem: users failed by the most accurate model likely have few options for alternative models that could work for them.

\subsection{Model improvement analysis is insensitive to threshold}
\label{threshold}
\begin{figure}
    \centering
    \includegraphics[width=\textwidth]{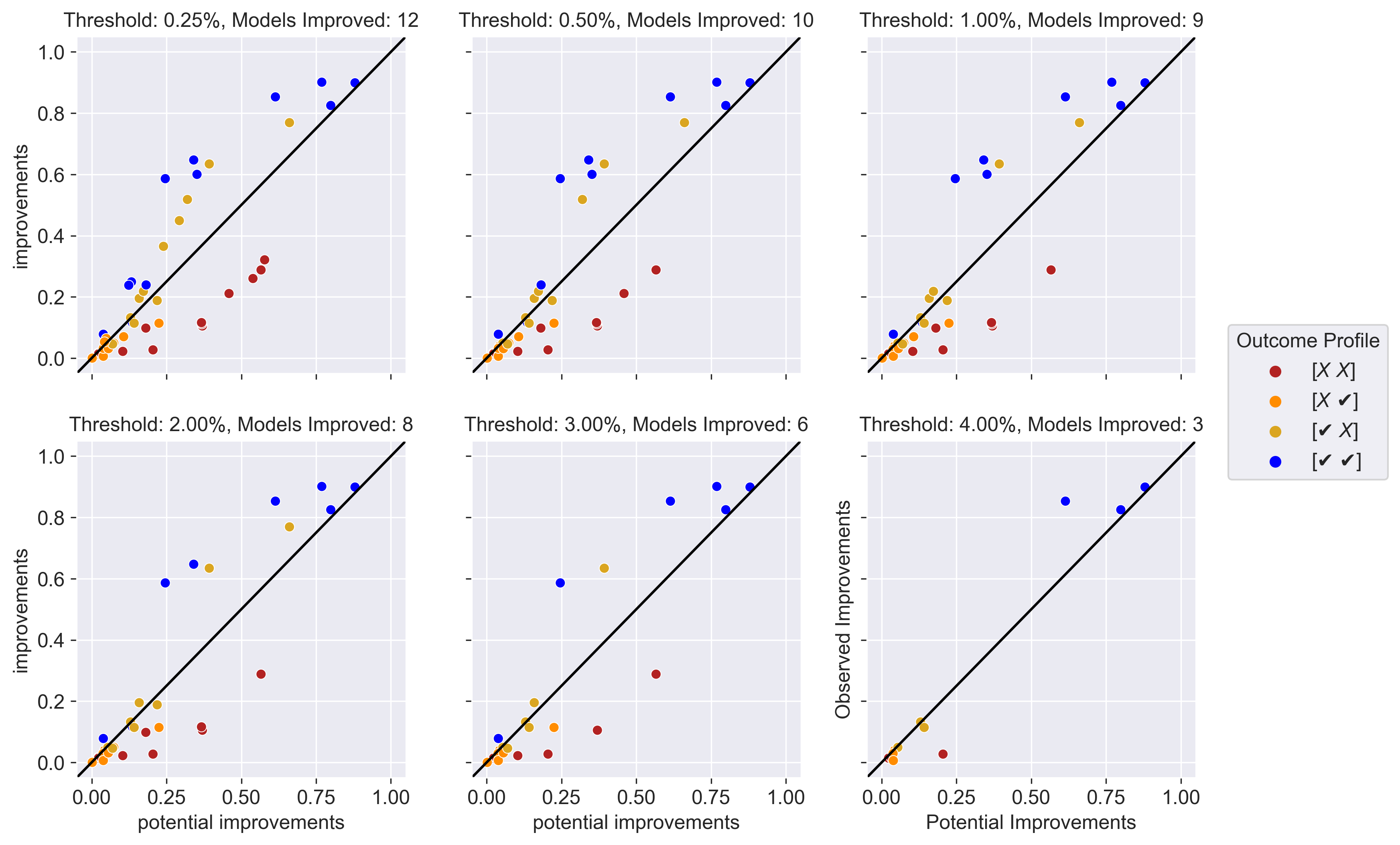}
    \caption{We replicate the graph from \autoref{fig:improvements} but with various thresholds for how much a model must improve for us to include it in the analysis. The patterns discussed in \autoref{sec:hapi-improvements} are consistent across choice of threshold.}
    \label{fig:improvements by threshold}
\end{figure}
In \autoref{sec:hapi-improvements}, we study how the improvement of a single model, in the sense that it becomes more impact, manifests at the \systemlevel.
To define improvement, we set the (slightly arbitrary) threshold that the model's accuracy improve by $0.5\%$, which we found to be large enough to be material to a model's performance while small enough to capture most model improvements. 
In \autoref{fig:improvements by threshold}, we plot the outcome profile distribution in the observed improvements set against the distribution over the potential improvements (as in \autoref{fig:all datasets improvements}) for 6 different thresholds of change. 
Across all thresholds, the patterns we discuss in \autoref{sec:hapi-improvements} hold: namely, models consistently under-improve on systemic failures.
This confirms that the findings and qualitative understanding we present is not particularly sensitive to the exact value of this threshold.

\subsection{Net Improvements}
\label{netimprovements}
\begin{figure}
    \begin{subfigure}[t][][t]{.5\textwidth}
        \centering
        \includegraphics[width=\textwidth]{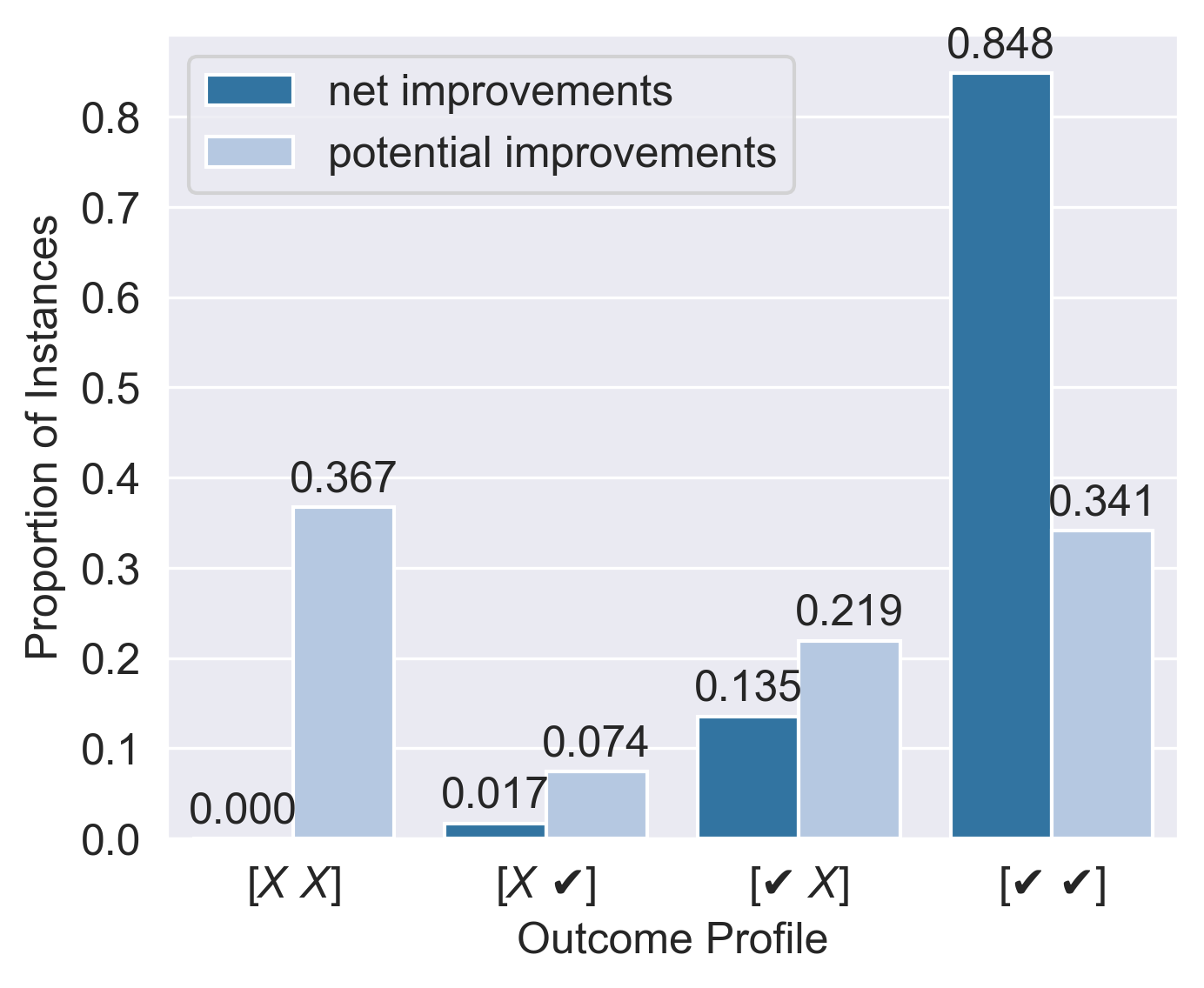}
        \caption{Outcome profile distribution using \textit{net improvements} instead of \textit{gross improvements} for (Baidu, Google) when Amazon improves on \waimai from 2020 to 2021.}
        \label{fig:net WAIMAI improvements}
    \end{subfigure}
    \hspace{1em}
    \begin{subfigure}[t]{.5\textwidth}
        \centering
        \includegraphics[width=\textwidth]{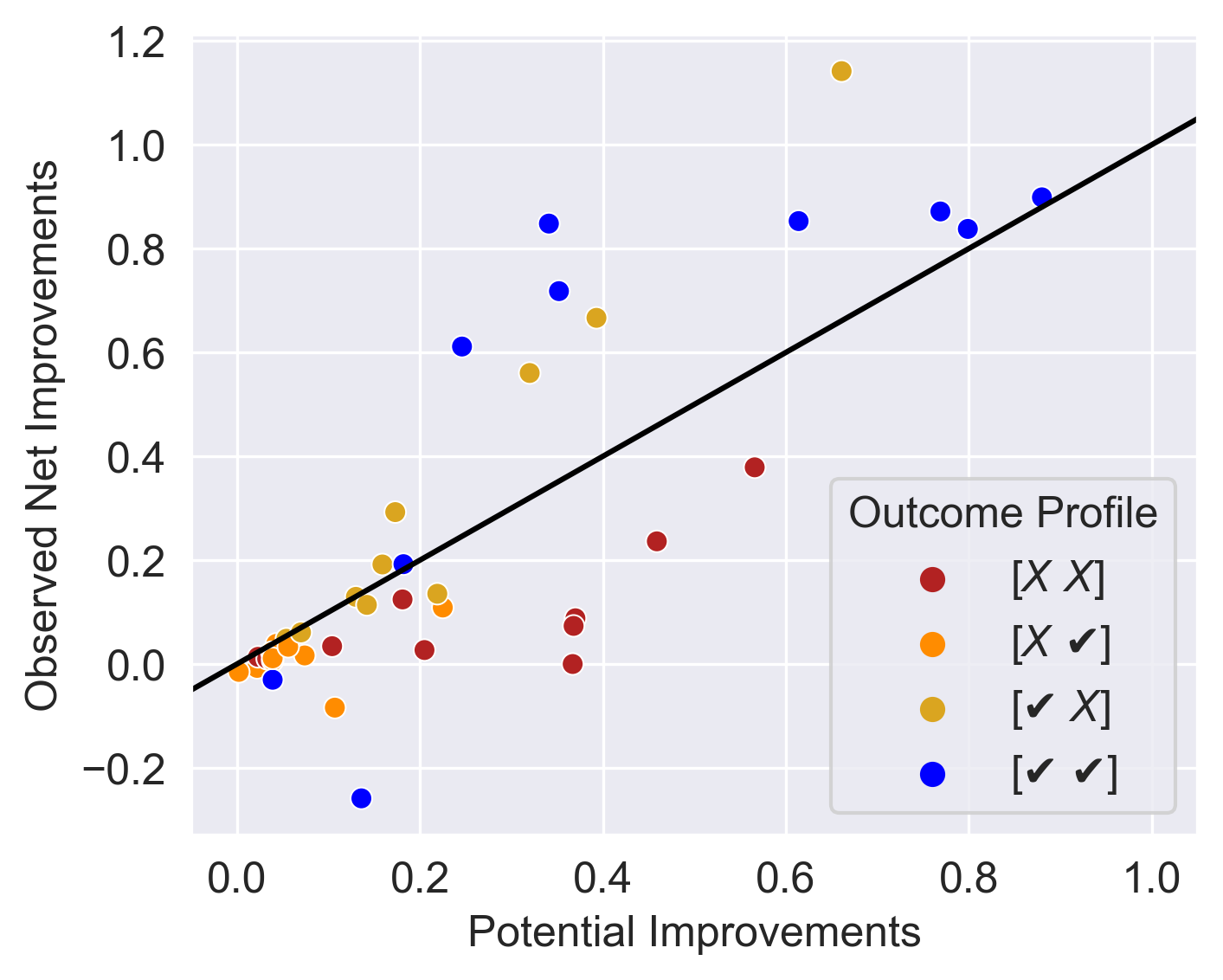}
        \caption{
        The distribution of outcome profiles for \textit{net improvements} instead of \textit{gross improvements} for all year-over-year model improvements across all datasets.}
        \label{fig:net improvements all datasets}
    \end{subfigure}
    \caption{
    We replicate the graphs in \autoref{fig:improvements} but using `net improvements' instead of `gross improvements'. The overall trends are consistent between the two experiments: namely, \himproved under-improves on systemic failures. 
    }
    \label{fig:net improvements}
\end{figure}
In \autoref{sec:hapi-improvements}, we define \textit{improvements} as the specific instances that \himproved misclassifies in the first year and correctly classifies in the second year. 
However, the size of this set --- hereafter, \textit{gross improvements} --- is always larger than the number of \textit{net improvements} the model makes because updates to the model tend to improve on some instances and regress on others.
That is, there are two competing forces when models change: the instances the model flips from incorrect to correct, but also the instances the models from correct to incorrect.
The difference of (i) and (ii) is the number of net improvements.

In \autoref{fig:net improvements}, we use net improvements instead of gross improvements, replicating the plot from in \autoref{fig:improvements}. 
To calculate the outcome profile distribution over net improvements, we subtract the number of gross declines from the number of gross improvements for each outcome profile: the denominator is the number of gross improvements minus the number of gross declines across all outcome profiles. 

The trends are generally consistent with what we observed when using gross improvements: \himproved tends to make little progress on systemic failures. 
In fact, when considering net improvements, models make even less improvement on systemic failures than before.
We highlight \waimai as a striking case study: there is no net improvements on systemic failures, despite a 2.5\% decrease in model error at the population level.

Overall, we emphasize that we recommend future research conducts analyses with both notions of improvements.
While we expect in many cases, as we have seen here, that the qualitative trends will be similar, the interpretations may differ.
For example, gross improvements more directly attend to the concern that there are some individuals who, year-over-year, continue to be failed by some or all models in the \system.
In contrast, net improvements more directly matches the sense in which models are improving.

\subsection{When Models Get Worse}
As a related question to what we examine in \autoref{fig:all datasets improvements}, we examine what happens to \systemlevel outcomes when a model gets worse. We find that, when models get worse, they disproportionately introduce new systemic failures into the system by 'over-declining' on instances that other models were already failing for. This further highlights how single-model measurements often fails to align with ecosystem-level outcomes.

\begin{figure}
    \centering
    \includegraphics{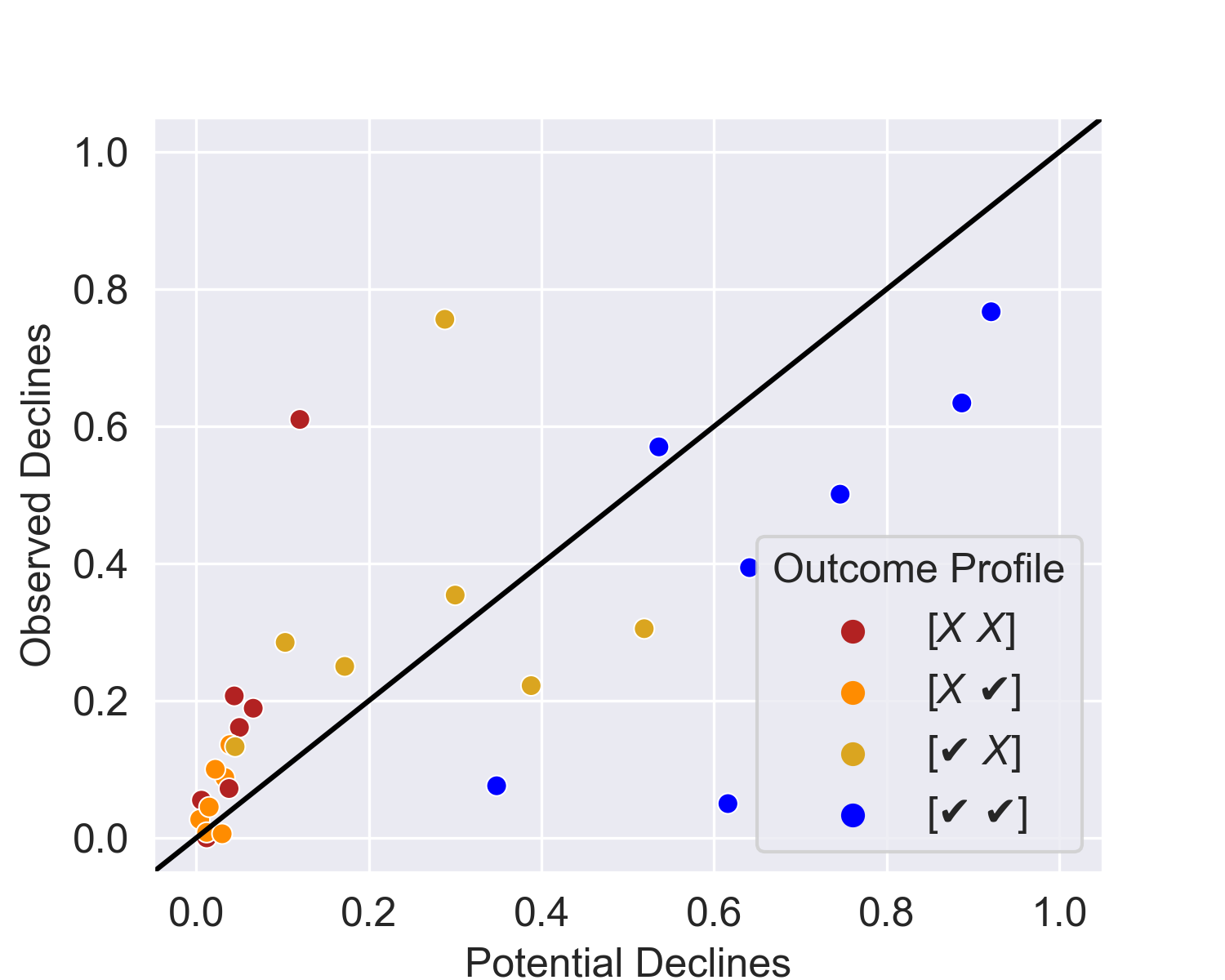}
    \caption{The distribution of outcome profiles for all year-over-
    year model declines across all datasets}
    \label{fig:enter-label}
\end{figure}

\section{Dermatology Experiments}
\label{dermatology}

\subsection{Data}

We work with \ddi (Diverse Dermatology Images), a dataset introduced by \citet{daneshjou2022disparities}, which contains predictions from 3 models and 2 board-certified dermatologists on 656 skin lesion images; the task is to predict whether a lesion is malignant or benign. The ground truth label comes from an external-source: in this case, a biopsy of the lesion, which is considered the gold-standard labeling procedure in this domain.

The 3 evaluated models include ModelDerm \citep{han2020augment}, a publicly available ML API, and two models from the academic literature -- DeepDerm \citep{esteva2017dermatologist} and HAM10k \citep{tschandl2018ham10000} -- that were chosen by \citet{daneshjou2022disparities} on the basis of their "popularity, availability, and previous demonstrations of state-of-the-art performance." Note that, in this case, none of the models have been trained on any portion of the \ddi dataset; the entire dataset serves as a test set.

In addition, each image is annotated with skintone metadata using the Fitzpatrick scale according to one of three categories: 
Fitzpatrick I \& II (light skin), Fitzpatrick III \& IV (medium skin), and Fitzpatrick V \& VI (dark skin). 
For all instances, the Fitzpatrick classification was determined using consensus review of two board-certified dermatologists. Additionally, a separate group of dermatologists rated the image quality of each image and discarded any low quality images; there was no significant difference in image quality ratings between images of different FST classifications.

Data on model and dermatologist predictions was graciously provided by \citet{daneshjou2022disparities}, subject to the terms of their standard research use agreement described in \url{https://ddi-dataset.github.io/index.html#access}.

\subsection{Analyses are insensitive to including/excluding HAM10k}
\begin{figure}
    \begin{subfigure}[t][][t]{.5\textwidth}
        \centering
        \includegraphics[width=\textwidth]{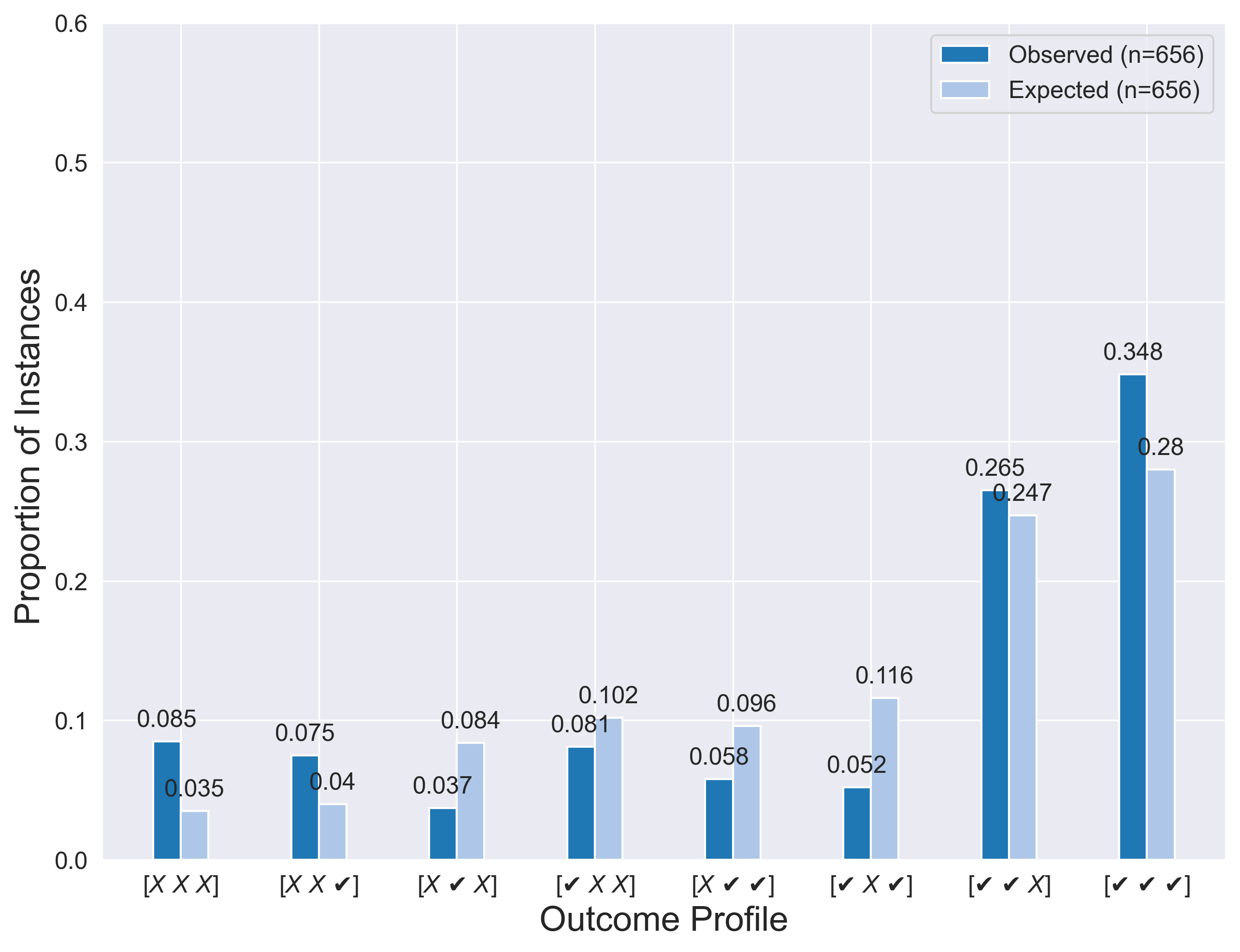}
        \caption{Model outcomes when including HAM10K yield even more \polarizedprofiles on \ddi than in \autoref{fig:human vs model polarization}}
        \label{fig:ham10k outcomes}
    \end{subfigure}
    \hspace{1em}
    \begin{subfigure}[t]{.5\textwidth}
        \centering
        \includegraphics[width=\textwidth]{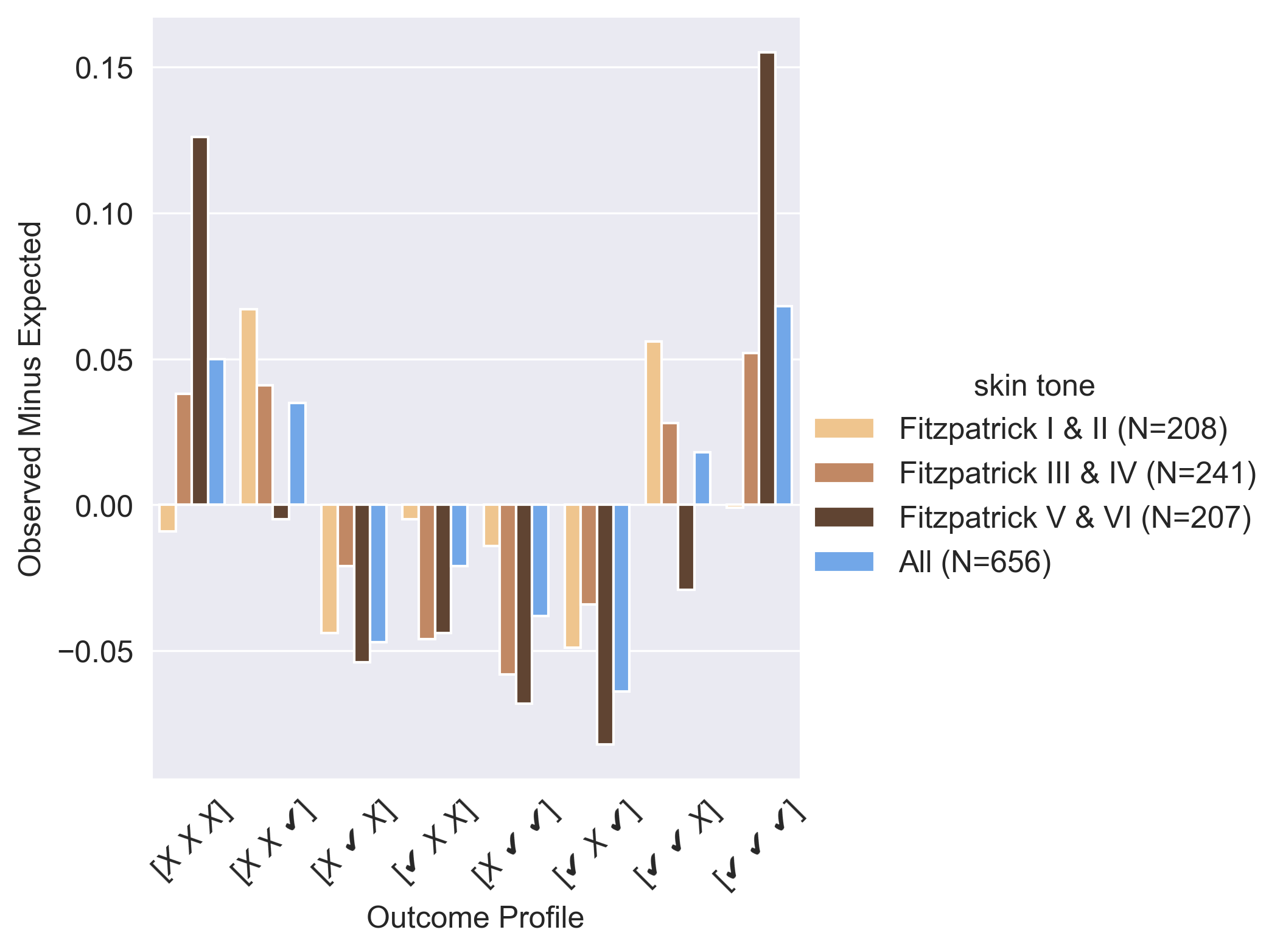}
        \caption{The inclusion of HAM10K yields more pronounced racial disparities than in \autoref{fig:derm_diff_by_race}
        .}
        \label{fig:ham10k diff by race}
    \end{subfigure}
    \caption{
    We replicate the graphs in \autoref{fig:human vs model polarization} and \autoref{fig:derm_diff_by_race} but with the inclusion of HAM10K. \Profilepolarization and racial disparities in models are even more pronounced when including HAM10k.
    }
    \label{fig:ham10k results}
\end{figure}
In \autoref{sec:dermatology}, we don't include predictions from Ham10K because the model predicts a negative on almost all instances: it has a precision of $0.99$ but a recall of $0.06$. 

We decided to remove HAM10k because the pattern of near-universal negative predictions does not reflect model behavior we would expect of models deployed in clinical settings.
Namely, even if deployed, the structure of the model errors are not particularly interesting and are largely predictable (in direction).
Beyond these fundamental reasons for excluding the model, we also removed the model for reasons unique to our analysis.
Including the model would have introduced an explicit class correlation in systemic failures (\ie almost all systemic failures would be malignant instances and none would be benign instances) and would have complicated the comparison with humans since there would be three models but only two humans.

To confirm that our findings hold independent of this choice, in \autoref{fig:ham10k results}, we replicate \autoref{fig:human vs model polarization} and \autoref{fig:derm_diff_by_race} but with outcomes from HAM10k included. 
We find that inclusion of HAM10k exacerbates the \profilepolarization of model outcomes and the racial disparities in model outcomes.

\end{document}